\documentclass{article}
\usepackage{main,times}

\usepackage{amsmath,amsfonts,bm}

\def\eqref#1{equation~\ref{#1}}

\def\1{\bm{1}}

\DeclareMathAlphabet{\mathsfit}{\encodingdefault}{\sfdefault}{m}{sl}
\SetMathAlphabet{\mathsfit}{bold}{\encodingdefault}{\sfdefault}{bx}{n}

\definecolor{paleRed}{RGB}{255,204,204}
\definecolor{palePurple}{RGB}{210,200,250}
\definecolor{paleBlue}{RGB}{195,215,236}
\usepackage{hyperref}
\usepackage{url}
\usepackage{booktabs}
\usepackage{subcaption}
\usepackage{wrapfig}
\usepackage{xcolor}
\usepackage{cleveref}
\usepackage{amssymb}
\usepackage{graphicx} 
\usepackage{listings}
\usepackage{enumitem}
\usepackage{adjustbox}
\usepackage{textpos}
\usepackage{orcidlink}
\usepackage{amsfonts}
\usepackage{url}
\usepackage{graphicx}
\usepackage{wrapfig}
\usepackage{tabularx}
\usepackage{color, colortbl}
\usepackage{multirow}
\usepackage{xcolor}
\usepackage{graphicx}
\usepackage{footnote}
\usepackage{tabularx}
\usepackage{lipsum}
\usepackage{multicol}
\usepackage{siunitx}
\usepackage{tabularx,booktabs}
\usepackage{etoolbox}
\usepackage{algorithm}
\usepackage{listings}
\usepackage{booktabs}
\usepackage{threeparttable}

\usepackage{etoolbox}
\makeatletter
\AfterEndEnvironment{algorithm}{\let\@algcomment\relax}
\AtEndEnvironment{algorithm}{\kern2pt\hrule\relax\vskip3pt\@algcomment}
\let\@algcomment\relax
\newcommand\algcomment[1]{\def\@algcomment{\footnotesize#1}}
\renewcommand\fs@ruled{\def\@fs@cfont{\bfseries}\let\@fs@capt\floatc@ruled
  \def\@fs@pre{\hrule height.8pt depth0pt \kern2pt}%
  \def\@fs@post{}%
  \def\@fs@mid{\kern2pt\hrule\kern2pt}%
  \let\@fs@iftopcapt\iftrue}
\makeatother

\definecolor{deemph}{gray}{0.6}

\definecolor{LightCyan}{rgb}{0.88,1,1}
\definecolor{LightRed}{rgb}{1,0.5,0.5}
\definecolor{LightYellow}{rgb}{1,1,0.88}
\definecolor{Grey}{rgb}{0.75,0.75,0.75}
\definecolor{DarkGrey}{rgb}{0.55,0.55,0.55}
\definecolor{DarkGreen}{rgb}{0,0.65,0}

\title{EchoGen: Generating Visual Echoes in Any Scene via Feed-Forward Subject-Driven Auto-Regressive Model}

\author{\textbf{Ruixiao Dong}$^{1, 2}\thanks{Equal contribution.}$ \quad \textbf{Zhendong Wang}$^{1 *}$ \quad  \textbf{Keli Liu}$^{1}$ \quad \textbf{Li Li}$^{1}\thanks{Corresponding authors.}$ \quad \textbf{Ying Chen}$^{2\dag}$ \quad \textbf{Kai Li}$^{2}$ \\ \textbf{Daowen Li}$^{2}$\quad \textbf{Houqiang Li}$^{1}$ \\
$^{\text{1}}$ University of Science and Technology of China \\
$^{\text{2}}$ Alibaba Group \\
    {\tt\small \{dongruixiaoyx,zhendongwang,sa23006063\}@mail.ustc.edu.cn} \\
    {\tt\small \{lil1,lihq\}@ustc.edu.cn} \\
    {\tt\small \{chenying.ailab,kaishi.lk,lidaowen.ldw\}@alibaba-inc.com}\\
}

\iclrfinalcopy
\begin{document}

\maketitle

\begin{figure}[h]
    \centering    \includegraphics[width=0.99\linewidth]{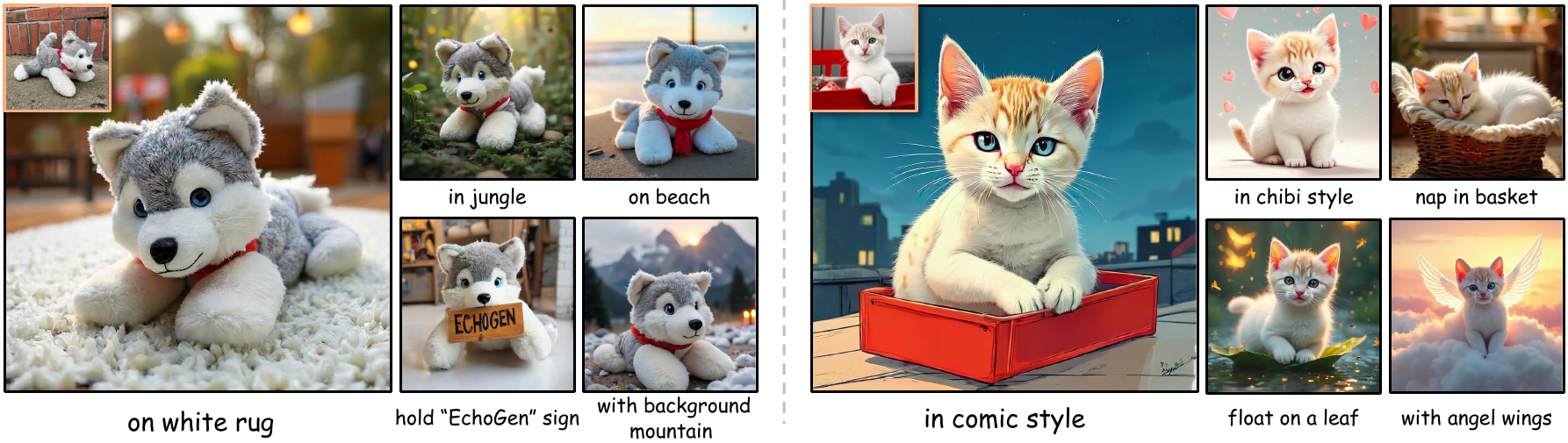}
    \caption{\textbf{Feed-forward subject-driven generation by EchoGen}. By employing a visual autoregressive paradigm, EchoGen achieves both high-quality image synthesis with lower latency, preserving intricate subject identity with exceptional efficiency.}
    \label{fig:teaser}
\end{figure}

\begin{abstract}

Subject-driven generation is a critical task in creative AI; yet current state-of-the-art methods present a stark trade-off. They either rely on computationally expensive, per-subject fine-tuning, sacrificing efficiency and zero-shot capability, or employ feed-forward architectures built on diffusion models, which are inherently plagued by slow inference speeds. 
Visual Auto-Regressive (VAR) models are renowned for their rapid sampling speeds and strong generative quality, making them an ideal yet underexplored foundation for resolving this tension.
To bridge this gap, we introduce \textbf{EchoGen}, 
a pioneering framework that empowers VAR models with subject-driven generation capabilities.
The core design of EchoGen is an effective dual-path injection strategy that disentangles a subject's high-level semantic identity from its low-level fine-grained details, enabling enhanced controllability and fidelity. 
We employ a semantic encoder to extract the subject's abstract identity, which is injected through decoupled cross-attention to guide the overall composition. Concurrently, a content encoder captures intricate visual details, which are integrated via a multi-modal attention mechanism to ensure high-fidelity texture and structural preservation.
To the best of our knowledge, EchoGen is the first feed-forward subject-driven framework built upon VAR models. Both quantitative and qualitative results substantiate our design, demonstrating that EchoGen achieves subject fidelity and image quality comparable to state-of-the-art diffusion-based methods with significantly lower sampling latency.
The code and models are publicly available at \url{https://github.com/drx-code/EchoGen}.

\end{abstract}

\section{Introduction}
The rapid evolution of text-to-image synthesis models~\citep{saharia2022photorealistic, rombach2022high, batifol2025flux, esser2024scaling} has catalyzed a variety of novel applications~\citep{zhang2023adding}, among which subject-driven generation stands out as an important task. This task aims to accurately depict a specified subject within diverse, user-defined scenes described through text prompts, while rigorously upholding the subject's core identity. The early approaches~\citep{ruiz2023dreambooth, gal2022image, kumari2023multi} introduced a test-time fine-tuning paradigm that optimizes a large pretrained model using a few images for each new subject. Although effective in preserving identity to some extent, this per-subject optimization process is computationally expensive, demanding at least hundreds of training iterations and substantial GPU resources, ultimately resulting in a distinct model checkpoint for each subject. These limitations significantly hinder the practicality and scalability of the test-time fine-tuning paradigm in real-world applications.

To improve efficiency and practicality, a new class of feed-forward approaches has recently emerged~\citep{li2023blip,pan2024kosmosg, ye2023ip, tan2025ominicontrol,shin2025large} based on diffusion models~\citep{rombach2022high, podell2024sdxl,batifol2025flux}. Instead of fine-tuning on a small set of images for each new subject, feed-forward approaches perform a single, large-scale supervised fine-tuning on a vast dataset composed of triplets~(text, reference image, target image). The model is trained to learn a generalizable mapping from a subject image to the snapshot version in the specified scene. The single process of pretraining enables zero-shot generation at inference time--a novel subject can be synthesized immediately without any subject-specific fine-tuning, significantly reducing the initial setup cost and decreasing generation latency by eliminating the need for test-time optimization. Nevertheless, these methods still inherit the computational demands of the underlying diffusion models due to the iterative denoising process.

Inspired by autoregressive generation in language models~\citep{radford2018improving, achiam2023gpt}, autoregressive visual generation~\citep{esser2021taming, ramesh2021zero, sun2024autoregressive} has emerged as a compelling alternative to diffusion models. Unlike diffusion's iterative denoising, autoregressive models synthesize content sequentially, token by token. This paradigm is further advanced by the Visual Autoregressive (VAR) model~\citep{tian2024visual, han2025infinity}, which employs a coarse-to-fine \textit{next-scale} generation strategy instead of traditional \textit{next-token} generation. It first generates tokens for the global composition and then renders fine-grained details, capturing a complete hierarchical representation from structure to texture. The novel paradigm allows VAR to achieve superior performance compared to traditional autoregressive models, outperforming top-tier diffusion models while offering faster inference speed. Despite the inherent suitability of the autoregressive paradigm for fine-grained conditioning, its potential for controllable generation, especially in the feed-forward, subject-driven context, remains largely untapped compared to the wealth of research on diffusion-based methods. This critical gap severely limits the practical applicability of VAR models, hindering their adoption in real-world scenarios where subject control is paramount.

In this work, we aim to bridge this gap by leveraging the inherent advantages of VAR to build an effective, scalable, and highly controllable system for subject-driven image synthesis. We propose \textit{EchoGen}, the first efficient \textbf{feed-forward} autoregressive framework that generates faithful visual renditions of a given subject in arbitrary scenes. At the core of EchoGen is a \textit{dual-path} injection mechanism that disentangles semantic features from fine-grained details. We inject high-level fine-grained semantic features extracted by a semantic encoder based on the pretrained vision foundation model~(DINOv2~\citep{oquab2024dinov}) into the decoupled cross-attention layers~\citep{kumari2023multi} to bring structural and stylistic coherence while avoiding drift in prompt following.To enable global semantic conditioning, we prepend the global semantic embedding extracted from DINOv2 as a prefix and subsequently infuse it via Adaptive LayerNorm, thereby steering the overall semantic generation. However, generating with semantic features alone often misses low-level details. To complement these features, a second pathway employs a pretrained content encoder (FLUX.1-dev VAE~\citep{batifol2025flux}) to extract fine-grained image features, which are incorporated via a multi-modal attention module, ensuring faithful reconstruction of local textures and details.To preserve the generative capabilities of the pretrained VAR model, we adopt a parameter-efficient fine-tuning strategy that freezes the backbone and only updates key components within the subject injection modules. Extensive quantitative and qualitative evaluations on DreamBench~\citep{ruiz2023dreambooth} benchmark and human evaluation demonstrate that EchoGen achieves subject fidelity, text alignment, and image quality comparable to and even exceeding state-of-the-art diffusion-based methods, while exhibiting lower sampling latency.

Our principal contributions can be summarized as follows: 
\begin{itemize}
\item We introduce EchoGen, the first feed-forward, efficient, subject-driven generation framework built upon a visual autoregressive model. This establishes a compelling new paradigm for controllable subject-driven synthesis beyond the dominant diffusion-based approaches.

\item We propose a novel dual-path injection strategy that disentangles the identity of a subject into high-level semantics and fine-grained details. By injecting these features through separate pathways within a parameter-efficiently tuned model, EchoGen achieves faithful subject representation across diverse scenes.

\item Extensive experiments demonstrate that EchoGen achieves subject fidelity, text alignment, and image quality that are competitive with or superior to state-of-the-art diffusion-based methods with much faster inference speed.
\end{itemize}

\section{Related Works}
\subsection{Autoregressive Image Generation}
Unlike diffusion-based methods that synthesize images via iterative denoising, the autoregressive paradigm models image distributions by sequentially predicting visual tokens conditioned on the preceding context. This approach evolves from inefficient and low quality early pixel-level methods~\citep{van2016conditional,salimans2017pixelcnn} to a dominant two-stage framework that first compresses images into discrete tokens and then models their distribution utilizing Transformer~\citep{esser2021taming}. This paradigm substantially improves generation fidelity and efficiency, underpinning advances in text-to-image synthesis~\citep{ramesh2021zero,yu2022scaling} and controllable generation~\citep{ControlAR}. Subsequent work further refines it by improving image tokenizers~\citep{yu2022vectorquantized,mentzer2024finite}, exploring continuous representations with diffusion modeling~\citep{li2024autoregressive,fan2025fluid}, or adapting large language models for visual generation~\citep{sun2024autoregressive,wu2024janus}. To mitigate structural degradation induced by the fixed raster-scan order, Visual Autoregressive (VAR) models~\citep{tian2024visual} introduce a hierarchical coarse-to-fine strategy that progressively refines fine-grained details by next-scale prediction. The following version Infinity~\citep{han2025infinity} extends the VAR model to text-to-image generation, achieving superior quality with significantly lower sampling latency than diffusion models. While existing works extend VAR to controllable generation~\citep{yao2024car,li2024controlvar,chung2025fine}, feed-forward subject-driven personalization remains underexplored, limiting the practical applicability of the VAR framework.

\subsection{Subject-driven Image Generation}

\textbf{Test-time fine-tuning methods}. Diffusion models~\citep{ho2020denoising,rombach2022high} have achieved remarkable success in high-fidelity text-to-image (T2I) synthesis~\citep{podell2024sdxl,esser2024scaling,batifol2025flux}. For subject-driven tasks, relying solely on text prompts is often insufficient to preserve the defining characteristics of specific subjects. To address this, pioneering methods~\citep{gal2022image,ruiz2023dreambooth,kumari2023multi} introduce customization by fine-tuning on a small set of reference images for each target subject. While these approaches can capture intricate details and deliver high fidelity to some extent, their dependence on per-subject optimization remains time-consuming and computationally demanding, which limits practical use.

\textbf{Feed-forward subject-driven approaches}. To overcome the efficiency limitations of per-subject optimization, feed-forward methods have been developed~\citep{wei2023elite,zeng2024jedi, patel2024eclipse,ma2024subject,wang2025msdiffusion}. These models are trained once to condition on subject features from vision encoders, enabling fast, zero-shot synthesis for novel subjects. Early works such as BLIP-Diffusion~\citep{li2023blip} jointly fine-tune the denoising network with multi-modal alignment modules while suffering from inadequate fidelity and image quality. To mitigate the high computational cost of full model tuning, parameter-efficient strategies~\citep{pan2024kosmosg, ye2023ip,tan2025ominicontrol,zhang2025easycontrol,wu2025less} incorporate lightweight modules such as LoRA~\citep{hu2022lora} or adapters. These modules inject reference features into the diffusion transformer, typically via attention mechanisms, while keeping most pretrained weights frozen. However, since these methods all rely on diffusion backbones, they inherit the substantial inference latency of the iterative denoising process, which constrains their practical deployment. 

\section{Preliminary of visual autoregressive modeling}

Autoregressive models~\citep{esser2021taming} reframe image synthesis as a sequential token prediction, under the \textit{next-token} prediction. The image is first tokenized into a discrete feature map using a visual tokenizer $\mathcal{E}$ and then flattened into a one-dimensional sequence, typically following the raster scan order. The model is then trained to predict each token $x_i$ given the preceding tokens $(x_1,...,x_{i-1})$ and the condition $c$, factorizing the sequence distribution as
$p(x_1,\ldots,x_N|c)=\prod_{i=1}^{N} p(x_i |x_1,\ldots,x_{i-1},c)$.
However, the vanilla next-token paradigm with fixed raster order induces structural degradation and insufficient modeling. 

Visual autoregressive modeling~\citep{tian2024visual} addresses the above issues by shifting the prediction paradigm from \textit{next-token} to \textit{next-scale}: instead of predicting one token at a time, it predicts entire token maps at \textit{progressively increasing resolutions}. The visual encoder $\mathcal{E}$ first maps an image $I$ to latent $F$, and then produces $K$ multi-scale token maps $(r_1,...r_K)$ with increasing resolutions $h_k\times w_k$ by applying a residual vector quantizer. A GPT-style Transformer begins from the generation of the 1$\times$1 map $r_1$ and autoregressively predicts each subsequent scale given prior scales and condition $c$, achieving generation from global structure to fine details, which is formulated as:

\begin{equation}
\setlength\abovedisplayskip{3pt}
p(r_1,\ldots,r_K | c)=\prod_{k=1}^{K} p(r_k|r_1,\ldots,r_{k-1},c).
\setlength\belowdisplayskip{3pt}
\end{equation}

This scale-wise coarse-to-fine paradigm is well suited for scalable text-to-image generation. The text-to-image generation model Infinity~\citep{han2025infinity} leverages bitwise quantization to expand the vocabulary size under the next-scale paradigm, reporting state-of-the-art performance with reduced sampling latency compared to diffusion baselines. In this paper, to bypass the cumbersome per-subject fine-tuning and the heavy computational cost during inference, we propose a novel feed-forward framework based on VAR models, featuring a single parameter-efficient fine-tuning phase.

\section{EchoGen}
\begin{figure*}[t]
    \centering
    \includegraphics[width=0.99\linewidth]{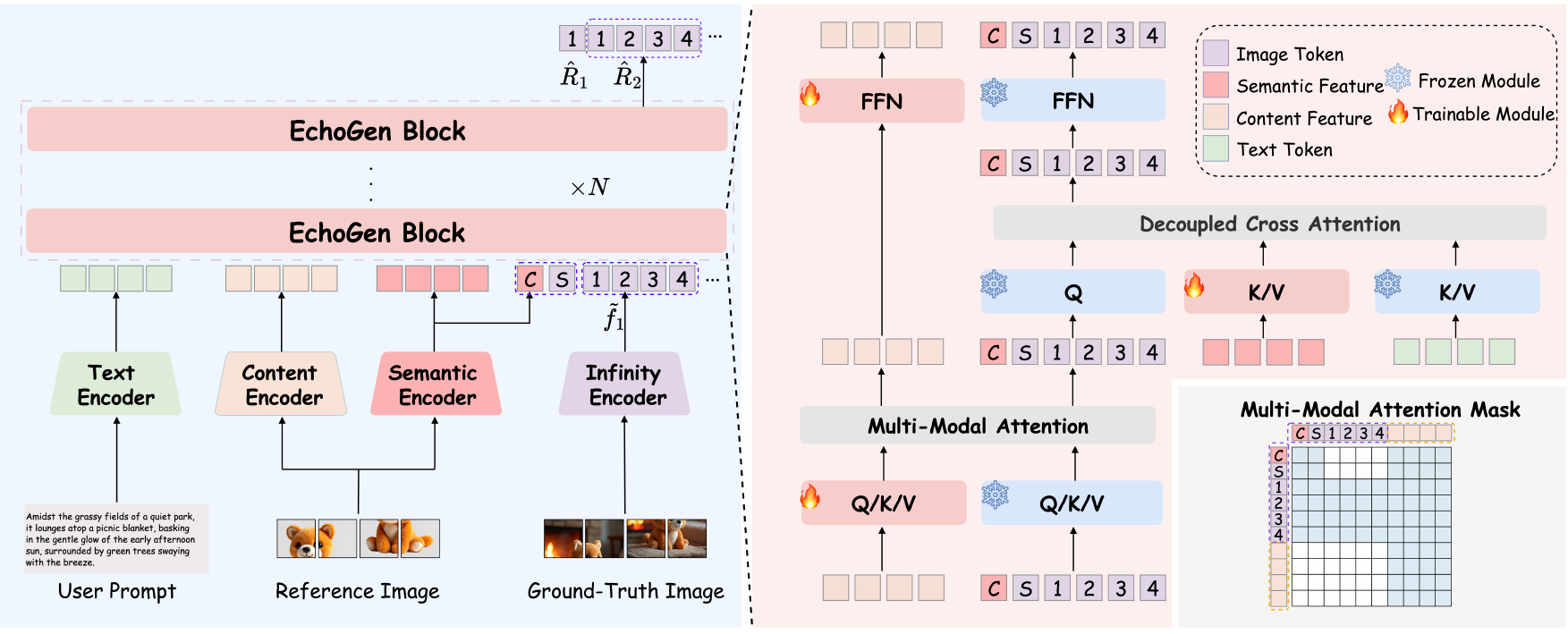}
    \caption{
    \textbf{Overview of the EchoGen architecture}. The left panel illustrates the overall model framework with dual-path subject injection, while the right panel provides a detailed schematic of the EchoGen block with a carefully designed attention mask applied in the Multi-Modal Attention module to avoid feature leakage. $C$ denotes the global semantic token extracted from the semantic encoder, which is prepended to the input sequence. $S$ represents the start token for the first-scale generation. Adaptive Layer Normalization modules in the EchoGen blocks are omitted for clarity.}
    \label{fig:framework}
\end{figure*}
\subsection{Overall Framework}
We are seeking a novel feed-forward framework for subject-driven generation built upon Infinity, based on the proposed \textbf{EchoGen} block with effective dual-path subject information injection, in which a content encoder and a semantic encoder cooperate to provide comprehensive subject features from both sides of a coin. The overview of the EchoGen architecture and its basic block is illustrated in \Cref{fig:framework}. Before subject injection, to ensure robustness against background noise that may interfere with subject injection, a pipeline based on the multi-modality model Qwen2.5-VL~\citep{Qwen2.5-VL} and the open segmentation model GroundingDINO~\citep{liu2024grounding} is carefully designed to segment the subject from complex scenes. Given the segmented subject image, our EchoGen model is trained using a parameter-efficient methodology that freezes the pretrained backbone while fine-tuning only newly introduced attention modules. During inference, we apply flexible subject-text classifier-free guidance for explicit control over the trade-off between subject fidelity and textual alignment, enabling versatile and controllable generation.

\subsection{Dual-Path Subject Injection}\label{sec:dual}

\textbf{Semantic feature injection for identity preservation}. 
The semantic feature, which captures abstract characteristics, provides a representation that is critical for avoiding the identity drift common in subject-driven generative models. Following this principle, we introduce a bifurcated injection strategy that targets both the fine-grained and global levels of the generative process. For fine-grained conditioning, we employ the pretrained DINOv2 vision encoder to extract patch-level semantic embeddings. These embeddings are synergistically integrated with the original textual conditioning via a decoupled cross-attention mechanism~\citep{kumari2023multi}. Our decoupled cross-attention mechanism operates on query features $\mathcal{Z}$, conditioning them on both the text embedding $\mathit{c_t}$ and the fine-grained semantic features $\mathit{c_s}$, formulated as follows:

\begin{equation}
\begin{aligned}
\mathcal{Q}&=\mathcal{Z}\mathit{W^q}, \, \mathcal{K}=\mathrm{concat}\left(\mathit{c_s} \mathit{W^k_s}, \mathit{c_t}\mathit{W^k_t}\right), \, \mathcal{V}=\mathrm{concat}\left(\mathit{c_s} \mathit{W^v_s}, \mathit{c_t} \mathit{W^v_t}\right), \\
\mathcal{Z'}&=\mathrm{Attention}\left(\mathcal{Q,K,V}\right)=\mathrm{Softmax}\left(\mathcal{QK}^\top/\sqrt{d}\right)\mathcal{V},
\end{aligned}
\end{equation}
where $\mathit{W^q}$ is the query projector, $(\mathit{W^k_t},\mathit{W^v_t})$ and $(\mathit{W^k_s},\mathit{W^v_s})$ are two distinct sets of $(k, v)$ projectors to embed text prompting $\mathit{c_t}$ and semantic injection $\mathit{c_s}$, respectively. 
The resulting key and value pairs for each condition are concatenated to form the final context vectors $\mathcal{K}$ and $\mathcal{V}$. 
We keep the projectors for text prompting $(\mathit{W^k_t},\mathit{W^v_t})$ and the query projector $\mathit{W^q}$ frozen while exclusively optimizing the key and value projectors $(\mathit{W^k_s},\mathit{W^v_s})$ that map the semantic features of the reference images, enabling an alignment mapping from the semantic visual space to the generator’s latent space without perturbing the pretrained knowledge.

Moreover, we prepend the DINOv2 global semantic token $C$ to the input sequence to impose holistic semantic guidance. At the same time, this global token also serves as a condition for the Adaptive Layer Normalization (AdaLN) layer in the proposed EchoGen block, following~\citep{han2025infinity}. The infusion of fine-grained and global semantics ensures comprehensive semantic-informed generation, promoting fine-grained fidelity and global structural coherence.

\textbf{Content feature injection for detail preservation}. 
While the semantic embeddings provide a robust identity preservation, their high abstraction leads to generation with insufficient subject details. To achieve high fidelity of the subject's content, we complement it with a content feature infusion mechanism. To be specific, EchoGen employs the FLUX.1-dev VAE to extract low-level content features $\mathit{c_c}$, which are then integrated via the multi-modal attention. The generation process is then steered by a carefully designed attention operation: generated tokens have unobstructed access to the reference tokens, allowing them to distill fine-grained visual cues on demand; conversely, a causal mask renders the reference tokens oblivious to the generated sequence, which is a critical constraint for ensuring the autoregressive sampling trajectory. This masking schema is precisely demonstrated in the lower-right inset of \Cref{fig:framework}.
Specifically, given the generated token sequence $\mathcal{Z}$ and the detailed content condition $\mathit{c_c}$, the multi-modal attention utilizes separate linear projections $(\mathit{W^q,W^k,W^v})$ for $\mathcal{Z}$ and $(\mathit{W^q_c,W^k_c,W^v_c})$ for the condition $\mathit{c_c}$, with the applied attention mask $\mathrm{Mask}$, and then calculate the generated sequence $\mathcal{Z}'$ and condition $\mathit{c_c}'$ via:

\begin{equation}
\begin{aligned}
\mathcal{Q}&=\mathrm{concat}\left(\mathcal{Z}\mathit{W^q},\mathit{c_c}\mathit{W^q_c}\right), \, \mathcal{K}=\mathrm{concat}\left(\mathcal{Z}\mathit{W^k}, \mathit{c_c}\mathit{W^k_c}\right), \, \mathcal{V}=\mathrm{concat}\left(\mathcal{Z}\mathit{W^v},\mathit{c_c}\mathit{W^v_c}\right), \\
\mathcal{Z'},c_c'&=\mathrm{Attention}\left(\mathcal{Q,K,V}, \mathrm{Mask}\right)=\mathrm{Softmax}\left(\mathrm{Mask}\left(\mathcal{QK}^\top/\sqrt{d}\right)\right)\mathcal{V}.
\end{aligned}
\end{equation}
The pathways for the generated token sequence remain frozen, while exclusively parallel attention projectors $(\mathit{W^q_c,W^k_c,W^v_c})$ and FFN modules for processing content features are optimized.

Through this dual-path subject injection strategy, our model faithfully preserves the salient visual characteristics of the reference image while simultaneously maintaining a strong adherence to the provided text instructions.

\subsection{Subject Segmentation}\label{sec:fe}
\begin{wrapfigure}{r}{0.5\textwidth}
\begin{center}
\vspace{-0.2cm}
\includegraphics[width=0.5\textwidth]{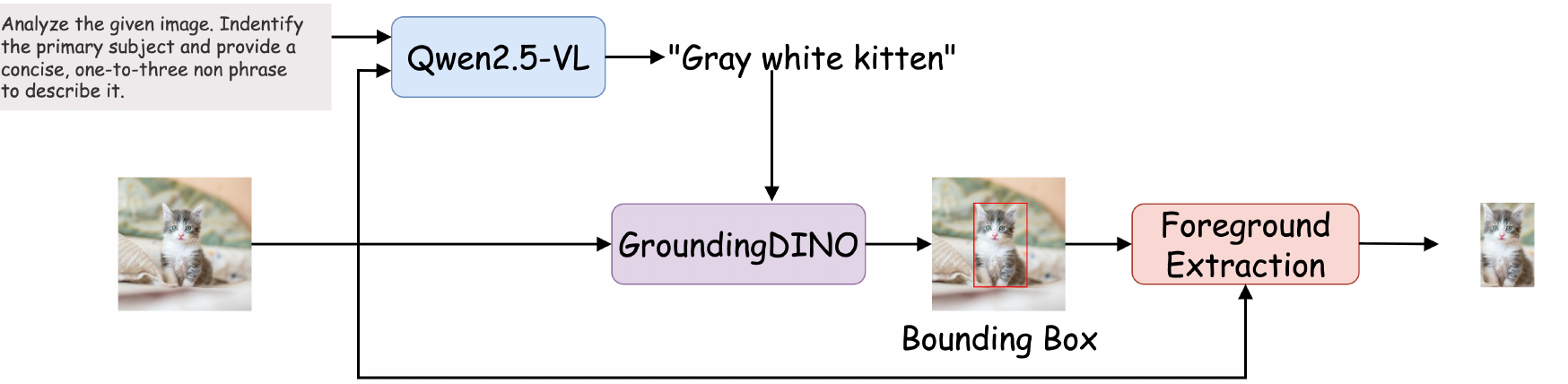}
\caption{\textbf{The pipeline of subject segmentation}.}
\label{fig:fe}
\end{center}
\vspace{-0.2cm}
\end{wrapfigure}
A common challenge in real-world scenarios is that user-provided reference images comprised of the subject of interest within visually complex backgrounds may harm the performance of subject injection. To mitigate this issue, we employ a subject segmentation pre-processing pipeline, illustrated in \Cref{fig:fe}. First, the Qwen2.5-VL~\citep{Qwen2.5-VL} vision-language model identifies the subject's semantic identity, producing a descriptive text prompt. This prompt is then used to condition the GroundingDINO~\citep{liu2024grounding} model for precise subject localization and bounding box generation. The foreground region is subsequently cropped according to this bounding box, while the surrounding unrelated regions are explicitly discarded and replaced with a uniform white background. This process ensures that subsequent feature injection operates attentively on the isolated representation of the referred subject.

\subsection{Sampling with Subject-Text Classifier-free Guidance} 
Classifier-Free Guidance (CFG)~\citep{ho2021classifierfree} has become a cornerstone technique for enhancing conditional control in generative models, especially in diffusion models. Its core principle is to amplify the conditional signal by extrapolating from an unconditional prediction towards a conditional one, thereby improving condition following at the cost of some diversity. 
Recently, many autoregressive models~\citep{chang2023muse,tian2024visual} have also incorporated CFG into their frameworks. In this work, we further enhance the influence of the text embedding $c_t$ and the subject condition $c_s,c_c$ within the CFG scheme for subject-driven generation. During training, we independently replace the text condition $c_t$ with an unconditional token $\varnothing_t$ and the image condition $c_s,c_c$ with unconditional embeddings $\varnothing_s,\varnothing_c$, each with a probability of 10\%. During inference, assuming the independence between the text condition $c_t$ and the image condition $c_s,c_c$, we compute the final logits predicted by EchoGen via a flexible guidance rule that integrates both controls:
\begin{equation}
\hat{l} = l(\varnothing_t, \varnothing_s,\varnothing_c) + \gamma_t\times(l(c_t, \varnothing_s,\varnothing_c) - l(\varnothing_t, \varnothing_s,\varnothing_c)) + \gamma_I \times(l(c_t, c_s,c_c) - l(c_t, \varnothing_s,\varnothing_c)),
\end{equation}
where $l$ denotes the Transformer output logits, and $\gamma_t$ together with $\gamma_I$ are hyperparameters that govern the guidance scales. This dynamic text-subject guidance not only strengthens the influence of text embeddings and image prompts, thereby improving generation performance, but also provides a flexible mechanism to balance text alignment with the reference preservation. 

\section{Experiment}

\subsection{Setup}

\textbf{Datasets}. We conduct experiments on a merged dataset curated from the Subjects200K~\citep{tan2025ominicontrol} and UNO-1M datasets~\citep{wu2025less}, yielding a large-scale high-quality corpus of approximately 640,000 triplets~(text prompts, reference images, and target images).
The corpora were synthetically generated using large language models (\textit{e.g.}, GPT-4o) and text-to-image generative models (\textit{e.g.}, FLUX.1-dev), and with image resolutions larger than 500$\times$500. For EchoGen-0.1B training, both the reference and target images are resized and center-cropped to 256$\times$256. To enable high-resolution generation for EchoGen-2B training, we avoid direct interpolation, which may introduce undesirable artifacts; instead, we upscale the images to 1024$\times$1024 using the PiSA-SR super-resolution model~\citep{sun2025pixel}. 

\textbf{Training details}. Our training protocol largely follows Infinity~\citep{han2025infinity}. We train EchoGen for 80K iterations, utilizing the AdamW~\citep{loshchilov2017decoupled} optimizer with a global batch size of 128, setting the base learning rate as $3\times 10^{-5}$ and the momentum parameters $(\beta_1, \beta_2) = (0.9, 0.97)$. To stabilize fine-tuning, we apply a reduced learning rate of $3\times 10^{-6}$ to the multi-modal attention parameters. More training details can be found in the appendix ~\ref{sec:traindetail}.

\begin{table*}
\centering
\vspace{-0.4cm}
\setlength{\tabcolsep}{3.6pt}
\begin{tabular}{llcccc}
\toprule
Method & Base Model & DINO$\uparrow$ & CLIP-I$\uparrow$ & CLIP-T$\uparrow$ & Latency $\downarrow$\\
\midrule
\multicolumn{6}{c}{\textcolor{gray}{\emph{Test-time Fine-tuning}}}\\
Textual-Inversion~\citep{gal2022image} & SD-v1.5 & 0.569 & 0.780 & 0.255 & 50min\\
DreamBooth~\citep{ruiz2023dreambooth} & SD-v1.5 & 0.668 & 0.803 & 0.305 & 15min\\
BLIP-Diffusion~\citep{li2023blip} & SD-v1.5 & 0.670 & 0.805 & 0.302 & -\\
AR-Booth~\citep{chung2025fine} &  Infinity-2B & \colorbox{palePurple}{\textbf{0.750}} & \colorbox{palePurple}{\textbf{0.808}} & 0.269 & 2.8h\\

\midrule
\multicolumn{6}{c}{\textcolor{gray}{\emph{Unified Generation}}}\\
OmniGen~\citep{xiao2025omnigen} & OmniGen & \colorbox{paleBlue}{\textbf{0.693}} & 0.801 & 0.315 & 93.4s\\

\midrule
\multicolumn{6}{c}{\textcolor{gray}{\emph{Feed-Forward }}}\\
ELITE~\citep{wei2023elite}  & SD-v1.4 & 0.621 & 0.771 & 0.293 & \colorbox{paleBlue}{\textbf{11.0s}}\\
Re-Imagen~\citep{chen2023reimagen}&Imagen& 0.600 & 0.740 & 0.270 &-\\
BLIP-Diffusion~\citep{li2023blip} & SD-v1.5 & 0.594 & 0.779  & 0.300 &- \\
$\lambda$-Eclipse~\citep{patel2024eclipse} & Kan-v2.2 & 0.613 & 0.783 & 0.307 & -\\
MS-Diffusion~\citep{wang2025msdiffusion} & SDXL & 0.671 & 0.792 & 0.321 & 39.6s\\
IP-Adapter~\citep{ye2023ip} & SDXL & 0.613 & \colorbox{palePurple}{\textbf{0.810}} & 0.292 & 16.9s\\
IP-Adapter~\citep{ye2023ip} & FLUX.1-dev & 0.561 & 0.725 & \colorbox{paleRed}{\textbf{0.351}} & -\\
OminiControl~\citep{tan2025ominicontrol} & FLUX.1-dev  &  0.684 & 0.799 & 0.312 & 27.5s\\
EasyControl~\citep{zhang2025easycontrol} & FLUX.1-dev & 0.652 & 0.789 & \colorbox{palePurple}{\textbf{0.325}} & 25.4s\\

\midrule
EchoGen-0.1B & Infinity-0.1B & 0.675 & 0.806 & 0.321 &\colorbox{paleRed}{\textbf{0.5s}}\\
EchoGen-2B & Infinity-2B &\colorbox{paleRed}{\textbf{0.755}} & \colorbox{paleRed}{\textbf{0.835}}& \colorbox{palePurple}{\textbf{0.325}}&\colorbox{palePurple}{\textbf{5.2s}}\\
\bottomrule
\end{tabular}
\caption{
\textbf{Quantitative comparisons on DreamBench~\citep{ruiz2023dreambooth}}. 
We highlight the \colorbox{paleRed}{\textbf{best}}, \colorbox{palePurple}{\textbf{second-best}}, and \colorbox{paleBlue}{\textbf{third-best}} values for each metric. The results indicate that EchoGen attains performance on par with diffusion-based approaches while delivering substantially faster sampling.
}
\label{tab:DreamBench}
\end{table*}

\begin{figure*}[t]
    \centering
    \vspace{-0.4cm}
    \includegraphics[width=0.99\linewidth]{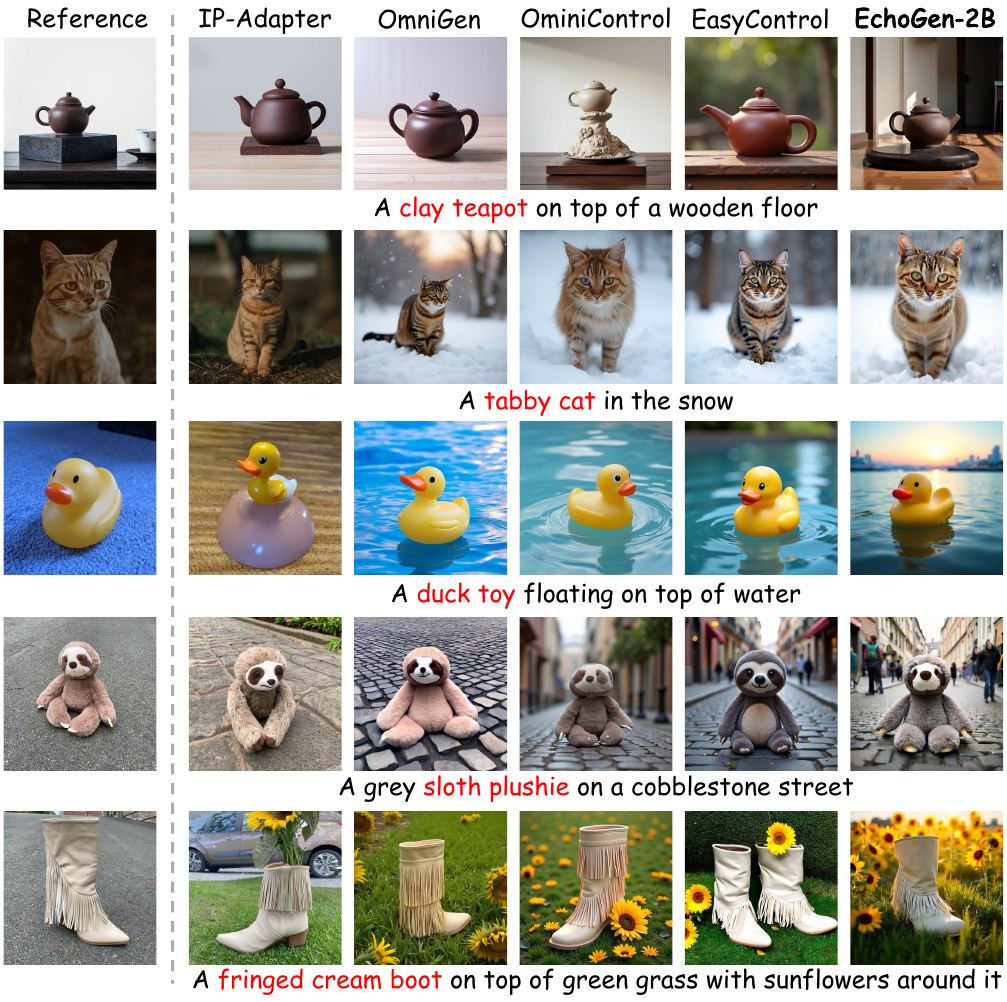}
    \caption{
    \textbf{Qualitative comparison with diffusion-based methods on DreamBench~\citep{ruiz2023dreambooth}}. For a fair comparison, we adopt the default sampling settings for all baseline models.}
    \label{fig:quali}
    \vspace{-0.2cm}
\end{figure*}

\begin{table*}[t]
\centering
\resizebox{1.0\textwidth}{!}{
\setlength{\tabcolsep}{6.5pt}
\begin{tabular}{lccc}
\toprule
Method & Subject Fidelity$\uparrow$ & Text Alignment$\uparrow$ & Photorealism$\uparrow$ \\
\midrule
OmniGen~\citep{xiao2025omnigen} & 0.15 & 0.13&  0.09  \\
IP-adapter~\citep{ye2023ip} & 0.21 &0.05 & 0.14\\
OminiControl~\citep{tan2025ominicontrol} & 0.12& 0.21& 0.15\\
EasyControl~\citep{zhang2025easycontrol} & 0.15& \textbf{0.31}& 0.28 \\
\midrule
EchoGen-2B & \textbf{0.37} & 0.30& \textbf{0.34}\\
\bottomrule
\end{tabular}
}
\caption{
\textbf{Human evaluation}. We compare our method with previous approaches based on three aspects: text alignment, subject fidelity, and photorealism. 
}
\label{tab:UserStudy}
\end{table*}

\textbf{Evaluation}. Following prior works~\citep{ruiz2023dreambooth,li2023blip}, we evaluate our approach in terms of subject fidelity and text alignment on the DreamBench benchmark~\citep{ruiz2023dreambooth}. Subject fidelity is measured by the cosine similarity between the generated and reference images using both CLIP~\citep{radford2021learning} image embeddings (CLIP-I) and DINO~\citep{zhang2022dino} features (DINO). Text alignment is assessed via the CLIP cosine similarity between the generated image and its corresponding input prompt (CLIP-T). DreamBench, comprising real-world images with prompt annotations, includes 30 unique subjects, each paired with 25 distinct prompts. Following the evaluation protocol of~\citep{pan2024kosmosg}, we select one reference image per subject, generate four images for each prompt–subject pair, yielding 3,000 generated images in total. DINO and CLIP-I scores are computed by comparing each generated image against its corresponding reference image. The sampling latency is measured on an H20 GPU for all methods.

\subsection{Main Results}

We compare EchoGen with three categories of prior works: (1) test-time fine-tuning methods that require per-subject optimization; (2) unified generation models with large-scale pre-training; and (3) feed-forward approaches that share the same paradigm as ours and constitute our most baselines.

\textbf{Quantitative results}. We benchmark EchoGen performance against contemporary subject-driven diffusion-based methods in the DreamBench dataset~\citep{ruiz2023dreambooth}, with quantitative results summarized in \Cref{tab:DreamBench}. EchoGen achieves performance that comparable or superior to leading diffusion-based approaches in the core metrics of subject fidelity and text alignment, and demonstrates balanced performance across evaluation axes. In contrast, several baselines, such as IP-Adapter~\citep{ye2023ip} exhibit significant weaknesses in specific metrics. Furthermore, the adoption of the visual autoregressive paradigm provides a clear efficiency advantage: EchoGen’s inference latency for a 1024$\times$1024 image is under 6 seconds, representing a significant acceleration over the more than 10 seconds required by its diffusion-based counterparts. 
Overall, these results indicate that EchoGen combines strong generative quality with markedly improved efficiency, offering a competitive alternative for subject-driven synthesis.
Since existing methods often adopt different evaluation protocols, we re-implement representative diffusion-based baselines and compare them with our model under a unified setting (see~\Cref{sec:more_abla}). The results demonstrate that our approach achieves comparable or superior performance while consistently reducing inference latency.

\textbf{Qualitative results}. \Cref{fig:quali} presents a rigorous qualitative comparison with prominent diffusion-based frameworks, revealing substantial advantages of our model in both subject fidelity and prompt correspondence. EchoGen exhibits the ability to render high-fidelity details, such as the precise reconstruction of the teapot spout and the nuanced texture of the sloth plushie, and we attribute this capability to our dual-path semantic-content feature injection design. In contrast, baselines including IP-Adapter~\citep{ye2023ip} and OminiControl~\citep{tan2025ominicontrol} exhibit characteristic failure cases, corroborating EchoGen’s robustness. EchoGen also demonstrates more consistent compliance with textual prompts, avoiding the language deviations observed in the generations of the duck toy and cat instances by IP-Adapter. 

\textbf{Human evaluation}. To assess the perceptual quality of EchoGen, we conduct a human evaluation study against strong baselines that span multiple categories of subject-driven methods. We focus on three criteria: text alignment, subject fidelity, and photorealism. The images are generated conditioned on the reference images and prompts sampled from DreamBench~\citep{ruiz2023dreambooth} and DreamBench++~\citep{peng2024dreambench++} benchmarks without cherry-picking, and for each criterion, participants select their preferred generated image among the outputs from five methods. We collect 450 responses from 25 participants, all with expertise in generative models, and report preference ratios in \Cref{tab:UserStudy}. The results show that EchoGen is preferred for subject fidelity and photorealism, surpassing all the diffusion-based contemporary baselines on these criteria. For text alignment, EchoGen performs on par with EasyControl~\citep{zhang2025easycontrol} and exhibits a clear advantage over the other compared methods.

\begin{figure*}[t]
    \centering
    \vspace{-0.5cm}
    \includegraphics[width=0.99\linewidth]{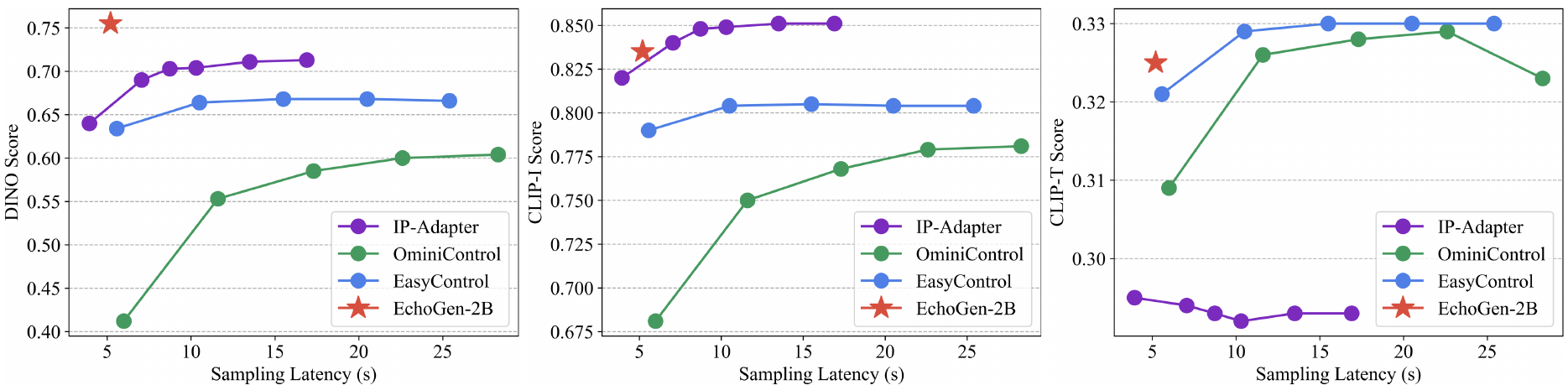}
    \caption{
    \textbf{Performance \textit{v.s.} sampling latency comparison among our EchoGen and baselines.}}
    \label{fig:samplingstep}
    \vspace{-0.6cm}
\end{figure*}

\textbf{Sampling Latency Analysis.}  We conduct a thorough analysis to evaluate the performance-latency trade-offs across all methods. Specifically, we re-produce diffusion-based methods with varying numbers of denoising steps in our evaluation protocol and report their performance versus sampling latency in \Cref{fig:samplingstep}. For the diffusion baselines, increasing the number of denoising steps improves subject fidelity (as measured by DINO, CLIP-I scores) up to a saturation point. In contrast, the text alignment (CLIP-T) score converges much earlier. Our model consistently offers a better trade-off, achieving comparable performance with significantly lower sampling latency than the diffusion-based baselines. This confirms the inherent efficiency and effectiveness of our approach.

\begin{wraptable}{r}{0.5\linewidth} 
    \centering
    \vspace{-0.4cm}
    \renewcommand{\arraystretch}{0.95}
    \setlength{\tabcolsep}{1.4pt}
    \begin{tabular}{cc}
    \toprule
    Model Component & Sampling Latency (s) \\
    \midrule
    Grounding-DINO & 0.24\\
    Semantic encoder & 0.01 \\
    Content encoder & 0.02 \\
    Infinity generator & 4.95 \\
    Qwen2.5-VL(Optional) & 1.13 \\
    \midrule
    EchoGen (w/o Qwen2.5-VL) & 5.22\\
    EchoGen (w/ Qwen2.5-VL) & 6.35 \\
    \bottomrule
    \end{tabular}
    \vspace{-0.2cm}
    \caption{\textbf{Per-component sampling latency measured on a single H20 GPU.}}
    \label{tab:latency}
    \vspace{-0.4cm}
\end{wraptable}
The detailed component-wise sampling latency of our EchoGen framework is provided in \Cref{tab:latency}. The results confirm that the framework's overall efficiency is not limited by auxiliary components such as Grounding-DINO. EchoGen maintains a significant speed advantage over diffusion-based methods, even with the inclusion of the optional Qwen2.5-VL model. Although this model is employed during training to automate subject identification for the GroundingDINO segmentation model, it is not required during inference. Instead, users can provide a descriptive text prompt (akin to the DreamBench format) for specifying the subject.

\subsection{Ablation Studies} 

We conduct a series of ablation studies to verify the effect of each component in EchoGen. Owing to computational constraints, all experiments are conducted on EchoGen-0.1B, using the same training settings to ensure fair ablation studies.

\begin{table}[t]
    \centering
    \setlength{\tabcolsep}{4pt}
    \begin{minipage}[t]{0.48\textwidth}
        \centering
        \begin{tabular}{lccc}
        \toprule
        Enc. & DINO$\uparrow$ & CLIP-I$\uparrow$ & CLIP-T$\uparrow$ \\
        \midrule
        SigLIP-2 & 0.438 & 0.720 & 0.320 \\
        FLUX.1-dev & 0.433 & 0.706 & 0.320 \\
        DINOv2 & \textbf{0.632} & \textbf{0.788} & \textbf{0.328}\\
        \bottomrule
        \end{tabular}
        \caption{
        \textbf{Significance of fine-grained semantic injection}. ``Enc.'' denotes the encoder type.
        }
        \label{tab:ca_cond}
    \end{minipage}
    \hfill 
    \begin{minipage}[t]{0.50\textwidth}
        \begin{tabular}{lccc}
        \toprule
        Exp. & DINO$\uparrow$ & CLIP-I$\uparrow$ & CLIP-T$\uparrow$ \\
        \midrule
        w/o prefix & 0.632 & 0.788 & \textbf{0.328}\\
        w prefix & \textbf{0.670} & \textbf{0.798} & 0.322 \\
        \bottomrule
        \end{tabular}
        \caption{
        \textbf{Ablation study on incorporating the global semantic features of reference images}. 
        }
        \label{tab:prefix}
    \end{minipage}
\end{table}

\textbf{Significance of fine-grained semantic information injection}. Fine-grained semantic conditional information is critical as it provides guidance for establishing the structure, enabling the model to synthesize stylistically and structurally coherent features consistent with the subject. Conversely, we argue that overly coarse-grained semantic features may fail to provide sufficient guidance for generating visually consistent echoes. To validate the importance of incorporating fine-grained semantic information, we conducted an ablation study with three distinct feature types independently injected via cross-attention:(1) coarse-grained semantic identity from SigLIP-2~\citep{tschannen2025siglip}, (2) fine-grained semantic features from DINOv2 and (3) FLUX.1-dev VAE features, which lack enough semantic information.
\Cref{tab:ca_cond} demonstrates that the fine-grained semantic DINOv2 features are the most suitable to represent the echo information in this task, as evidenced by all criteria. The failure of the SigLIP-2 and FLUX.1-dev VAE features can be attributed to their respective limitations: the former relies on features that are too coarse to guide subject generation, while the latter lacks semantic information.

\textbf{Incorporating the global semantic information}. To provide stronger guidance when constructing the global structure in the generation process, we prefix the global semantic $C$ token extracted by the DINOv2 semantic encoder and incorporate this token into the model via Adaptive LayerNorm. As demonstrated in \Cref{tab:prefix}, compared with solely relying on injecting fine-grained semantic features via cross-attention, the introduction of the global semantic token yields a substantial performance gain in consistency of subject characteristics, validating the effectiveness of incorporating global semantic information.

\begin{wraptable}{r}{0.5\linewidth} 
    \centering
    \setlength{\tabcolsep}{3.6pt}
     \vspace{-0.4cm}
    \begin{tabular}[t]{lccc}
    \toprule
    Module & DINO$\uparrow$ & CLIP-I$\uparrow$ & CLIP-T$\uparrow$ \\
    \midrule
    MM-Attn & 0.646 & 0.792 & \textbf{0.325} \\
    Cross-Attn & \textbf{0.670} & \textbf{0.798} & 0.322\\
    \bottomrule
    \end{tabular}
    \caption{
    \textbf{Different methods for incorporating semantic features}. 
    }
    \vspace{-0.2cm}
    \label{tab:dino_inject}
\end{wraptable}
\textbf{Distinct semantic feature injection strategies}. We explore the most effective method to guide the synthesis process conditioned on the semantic features of reference images. \Cref{tab:dino_inject} presents an analysis comparing two distinct feature injection modules: multi-modal attention and cross-attention. Our results indicate that while the multi-modal module achieves slightly better alignment with text prompts, the cross-attention mechanism yields significantly superior subject fidelity, as evidenced by a notably higher DINO score. Based on this, we opted to utilize cross-attention for injecting the semantic features in all subsequent experiments, rather than the multi-modal attention.

\begin{wraptable}{r}{0.5\linewidth} 
    \centering
    \setlength{\tabcolsep}{3.6pt}
     \vspace{-0.4cm}
    \begin{tabular}[t]{lccc}
    \toprule
    Exp. & DINO$\uparrow$ & CLIP-I$\uparrow$ & CLIP-T$\uparrow$ \\
    \midrule
    baseline & 0.670 & 0.798 & \textbf{0.322} \\
    +Cross-Attn & 0.667 & 0.803 & 0.318\\
    +MM-Attn & \textbf{0.672} & \textbf{0.806} & 0.321\\
    \bottomrule
    \end{tabular}
    \caption{
    \textbf{Impact of injecting subject details.}
    }
    \vspace{-0.2cm}
    \label{tab:content}
\end{wraptable}
\textbf{Enhancing subject fidelity with detailed content features.} Considering the absence of local details in the semantic features of the reference images, we incorporate a secondary pathway that injects localized content features of the subject. These features, extracted by the FLUX.1-dev VAE, are used to guide the synthesis of the fine-grained local details of the subject. The ablation study detailed in \Cref{tab:content} shows that employing a multi-modal attention mechanism to infuse these content features substantially improves the subject-fidelity of the generated samples, yielding a significant increase in CLIP-I. 

\begin{wrapfigure}{r}{0.55\textwidth}
    \centering
     \vspace{-0.4cm}
    \includegraphics[width=0.8\linewidth]{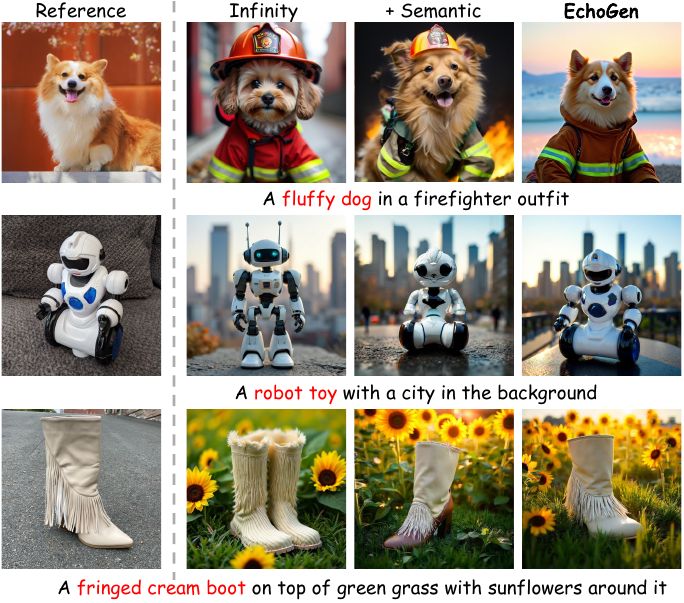}
    \caption{
    \textbf{Qualitative analysis of the effect of semantic and content feature injection.}}
    \label{fig:quali_abla}
    \vspace{-0.2cm}
\end{wrapfigure}
\noindent\textbf{Qualitative analysis of the effect of semantic and content feature injection.} We further qualitatively dissect the effect of each feature component to validate our design. As shown in \Cref{fig:quali_abla}, starting from the base Infinity backbone, introducing semantic features extracted by DINOv2 enables the generator to synthesize subjects that faithfully preserve the reference subject's structure and style. Moreover, further incorporating content features from the FLUX.1-dev VAE significantly enhances EchoGen's capability to render fine-grained, coherent details (\textit{e.g.}, the facial features of the robot toy and the fluffy dog, as well as the material and color of the shoe uppers). These qualitative results confirm the effectiveness of our dual-path injection design, where semantic and content features play distinct yet complementary roles.

\begin{wraptable}{r}{0.5\textwidth}
\centering
\vspace{-0.4cm}
\begin{tabular}{lccc}
\toprule
Exp. & DINO$\uparrow$ & CLIP-I$\uparrow$ & CLIP-T$\uparrow$ \\
\midrule
w/o SS  & 0.663 & 0.796 & 0.321 \\
w/ SS & \textbf{0.672} & \textbf{0.806} & 0.321 \\
\bottomrule
\end{tabular}
\caption{
\textbf{Enhancement by subject segmentation (denoted by SS) to mitigate background noise}.
}
\vspace{-0.2cm}
\label{tab:grounding}
\end{wraptable}
\textbf{Subject Segmentation}. To mitigate the influence of irrelevant background noise and focus on the primary subject, we leverage the Qwen2.5-VL vision language model~\citep{Qwen2.5-VL} and the GroundingDINO segmentation model~\citep{liu2024grounding} to segment the subject from the reference image. We conducted an ablation study, detailed in \Cref{tab:grounding}, to validate the efficacy of the echo segmentation protocol. The results confirm that the introduction of subject segmentation significantly enhances the generation performance, which is observed in the preservation of subject features, and demonstrate that isolating the main subject is critical to producing more accurate and faithful outputs.

\begin{figure*}[t]
    \centering
    \vspace{-0.4cm}
    \includegraphics[width=0.99\linewidth]{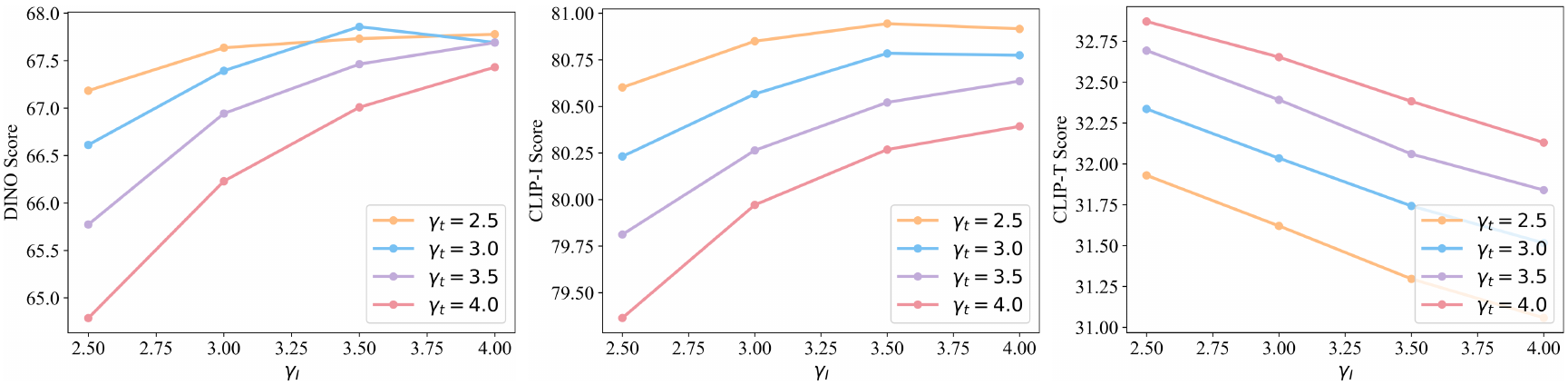}
    \caption{\textbf{Visualization of the effect of classifier-free guidance scale coefficient}.}
    \label{fig:cfg}
    \vspace{-0.4cm}
\end{figure*}
\textbf{Subject-text classifier-free guidance}. As detailed in \Cref{fig:cfg}, our experiments reveal a clear trade-off governed by the CFG hyperparameters within a proper scope. As the subject guidance weight $\gamma_I$ increases, subject fidelity improves, as indicated by higher CLIP-I and DINO scores. Conversely, this gain is accompanied by reduced text alignment, reflected in lower CLIP-T. The inverse relationship is observed when increasing the text condition scaling coefficient $\gamma_t$. This empirical result demonstrates the efficacy and flexibility of our CFG design, enabling users to dynamically adjust the balance between preserving reference image features and adhering to the text prompt.

\section{Conclusion}
This paper presents EchoGen, a novel framework for efficient, feed-forward subject-driven image synthesis based on a visual autoregressive paradigm, aiming to inherit the properties of high-quality generation and fast inference speed. Central to our methodology is a dual-path injection mechanism, meticulously designed to integrate both the semantic attributes and the precise textural details of reference images. Comprehensive evaluations corroborate the superiority of our design, revealing that EchoGen achieves generative performance on par with leading diffusion models while exhibiting substantially lower sampling latency. By pioneering a feed-forward, autoregressive solution for subject-driven synthesis, this research charts a new trajectory for the future development and application of visual autoregressive generative models.

\bibliography{main}

@article{radford2018improving,
  title={Improving language understanding by generative pre-training},
  author={Radford, Alec and Narasimhan, Karthik and Salimans, Tim and Sutskever, Ilya and others},
    year={2018}
}

@article{achiam2023gpt,
  title={Gpt-4 technical report},
  author={Achiam, Josh and Adler, Steven and Agarwal, Sandhini and Ahmad, Lama and Akkaya, Ilge and Aleman, Florencia Leoni and Almeida, Diogo and Altenschmidt, Janko and Altman, Sam and Anadkat, Shyamal and others},
  journal={arXiv preprint arXiv:2303.08774},
  year={2023}
}

@inproceedings{esser2021taming,
  title={Taming transformers for high-resolution image synthesis},
  author={Esser, Patrick and Rombach, Robin and Ommer, Bjorn},
  booktitle={Proceedings of the IEEE/CVF conference on computer vision and pattern recognition},
  pages={12873--12883},
  year={2021}
}

@article{sun2024autoregressive,
  title={Autoregressive model beats diffusion: Llama for scalable image generation},
  author={Sun, Peize and Jiang, Yi and Chen, Shoufa and Zhang, Shilong and Peng, Bingyue and Luo, Ping and Yuan, Zehuan},
  journal={arXiv preprint arXiv:2406.06525},
  year={2024}
}

@inproceedings{ramesh2021zero,
  title={Zero-shot text-to-image generation},
  author={Ramesh, Aditya and Pavlov, Mikhail and Goh, Gabriel and Gray, Scott and Voss, Chelsea and Radford, Alec and Chen, Mark and Sutskever, Ilya},
  booktitle={International conference on machine learning},
  pages={8821--8831},
  year={2021},
  organization={Pmlr}
}

@inproceedings{rombach2022high,
  title={High-resolution image synthesis with latent diffusion models},
  author={Rombach, Robin and Blattmann, Andreas and Lorenz, Dominik and Esser, Patrick and Ommer, Bj{\"o}rn},
  booktitle={Proceedings of the IEEE/CVF conference on computer vision and pattern recognition},
  pages={10684--10695},
  year={2022}
}

@article{ho2020denoising,
  title={Denoising diffusion probabilistic models},
  author={Ho, Jonathan and Jain, Ajay and Abbeel, Pieter},
  journal={Advances in neural information processing systems},
  volume={33},
  pages={6840--6851},
  year={2020}
}

@article{tian2024visual,
  title={Visual autoregressive modeling: Scalable image generation via next-scale prediction},
  author={Tian, Keyu and Jiang, Yi and Yuan, Zehuan and Peng, Bingyue and Wang, Liwei},
  journal={Advances in neural information processing systems},
  volume={37},
  pages={84839--84865},
  year={2024}
}

@inproceedings{han2025infinity,
  title={Infinity: Scaling bitwise autoregressive modeling for high-resolution image synthesis},
  author={Han, Jian and Liu, Jinlai and Jiang, Yi and Yan, Bin and Zhang, Yuqi and Yuan, Zehuan and Peng, Bingyue and Liu, Xiaobing},
  booktitle={Proceedings of the Computer Vision and Pattern Recognition Conference},
  pages={15733--15744},
  year={2025}
}

@inproceedings{ruiz2023dreambooth,
  title={Dreambooth: Fine tuning text-to-image diffusion models for subject-driven generation},
  author={Ruiz, Nataniel and Li, Yuanzhen and Jampani, Varun and Pritch, Yael and Rubinstein, Michael and Aberman, Kfir},
  booktitle={Proceedings of the IEEE/CVF conference on computer vision and pattern recognition},
  pages={22500--22510},
  year={2023}
}

@article{gal2022image,
  title={An image is worth one word: Personalizing text-to-image generation using textual inversion},
  author={Gal, Rinon and Alaluf, Yuval and Atzmon, Yuval and Patashnik, Or and Bermano, Amit H and Chechik, Gal and Cohen-Or, Daniel},
  journal={arXiv preprint arXiv:2208.01618},
  year={2022}
}

@inproceedings{kumari2023multi,
  title={Multi-concept customization of text-to-image diffusion},
  author={Kumari, Nupur and Zhang, Bingliang and Zhang, Richard and Shechtman, Eli and Zhu, Jun-Yan},
  booktitle={Proceedings of the IEEE/CVF conference on computer vision and pattern recognition},
  pages={1931--1941},
  year={2023}
}

@article{batifol2025flux,
  title={FLUX. 1 Kontext: Flow Matching for In-Context Image Generation and Editing in Latent Space},
  author={Batifol, Stephen and Blattmann, Andreas and Boesel, Frederic and Consul, Saksham and Diagne, Cyril and Dockhorn, Tim and English, Jack and English, Zion and Esser, Patrick and Kulal, Sumith and others},
  journal={arXiv e-prints},
  pages={arXiv--2506},
  year={2025}
}

@inproceedings{esser2024scaling,
  title={Scaling rectified flow transformers for high-resolution image synthesis},
  author={Esser, Patrick and Kulal, Sumith and Blattmann, Andreas and Entezari, Rahim and M{\"u}ller, Jonas and Saini, Harry and Levi, Yam and Lorenz, Dominik and Sauer, Axel and Boesel, Frederic and others},
  booktitle={Forty-first international conference on machine learning},
  year={2024}
}

@article{wu2025less,
  title={Less-to-more generalization: Unlocking more controllability by in-context generation},
  author={Wu, Shaojin and Huang, Mengqi and Wu, Wenxu and Cheng, Yufeng and Ding, Fei and He, Qian},
  journal={arXiv preprint arXiv:2504.02160},
  year={2025}
}

@article{li2023blip,
  title={Blip-diffusion: Pre-trained subject representation for controllable text-to-image generation and editing},
  author={Li, Dongxu and Li, Junnan and Hoi, Steven},
  journal={Advances in Neural Information Processing Systems},
  volume={36},
  pages={30146--30166},
  year={2023}
}

@article{ye2023ip,
  title={Ip-adapter: Text compatible image prompt adapter for text-to-image diffusion models},
  author={Ye, Hu and Zhang, Jun and Liu, Sibo and Han, Xiao and Yang, Wei},
  journal={arXiv preprint arXiv:2308.06721},
  year={2023}
}

@article{saharia2022photorealistic,
  title={Photorealistic text-to-image diffusion models with deep language understanding},
  author={Saharia, Chitwan and Chan, William and Saxena, Saurabh and Li, Lala and Whang, Jay and Denton, Emily L and Ghasemipour, Kamyar and Gontijo Lopes, Raphael and Karagol Ayan, Burcu and Salimans, Tim and others},
  journal={Advances in neural information processing systems},
  volume={35},
  pages={36479--36494},
  year={2022}
}

@inproceedings{zhang2023adding,
  title={Adding conditional control to text-to-image diffusion models},
  author={Zhang, Lvmin and Rao, Anyi and Agrawala, Maneesh},
  booktitle={Proceedings of the IEEE/CVF international conference on computer vision},
  pages={3836--3847},
  year={2023}
}

@inproceedings{tan2025ominicontrol,
  title={OminiControl: Minimal and Universal Control for Diffusion Transformer},
  author={Tan, Zhenxiong and Liu, Songhua and Yang, Xingyi and Xue, Qiaochu and Wang, Xinchao},
  booktitle={Proceedings of the IEEE/CVF International Conference on Computer Vision},
  year={2025}
}

@inproceedings{sun2025pixel,
  title={Pixel-level and semantic-level adjustable super-resolution: A dual-lora approach},
  author={Sun, Lingchen and Wu, Rongyuan and Ma, Zhiyuan and Liu, Shuaizheng and Yi, Qiaosi and Zhang, Lei},
  booktitle={Proceedings of the Computer Vision and Pattern Recognition Conference},
  pages={2333--2343},
  year={2025}
}

@article{loshchilov2017decoupled,
  title={Decoupled weight decay regularization},
  author={Loshchilov, Ilya and Hutter, Frank},
  journal={arXiv preprint arXiv:1711.05101},
  year={2017}
}

@inproceedings{radford2021learning,
  title={Learning transferable visual models from natural language supervision},
  author={Radford, Alec and Kim, Jong Wook and Hallacy, Chris and Ramesh, Aditya and Goh, Gabriel and Agarwal, Sandhini and Sastry, Girish and Askell, Amanda and Mishkin, Pamela and Clark, Jack and others},
  booktitle={International conference on machine learning},
  pages={8748--8763},
  year={2021},
  organization={PmLR}
}

@article{zhang2022dino,
  title={Dino: Detr with improved denoising anchor boxes for end-to-end object detection},
  author={Zhang, Hao and Li, Feng and Liu, Shilong and Zhang, Lei and Su, Hang and Zhu, Jun and Ni, Lionel M and Shum, Heung-Yeung},
  journal={arXiv preprint arXiv:2203.03605},
  year={2022}
}

@article{peng2024dreambench++,
  title={Dreambench++: A human-aligned benchmark for personalized image generation},
  author={Peng, Yuang and Cui, Yuxin and Tang, Haomiao and Qi, Zekun and Dong, Runpei and Bai, Jing and Han, Chunrui and Ge, Zheng and Zhang, Xiangyu and Xia, Shu-Tao},
  journal={arXiv preprint arXiv:2406.16855},
  year={2024}
}

@inproceedings{xiao2025omnigen,
  title={Omnigen: Unified image generation},
  author={Xiao, Shitao and Wang, Yueze and Zhou, Junjie and Yuan, Huaying and Xing, Xingrun and Yan, Ruiran and Li, Chaofan and Wang, Shuting and Huang, Tiejun and Liu, Zheng},
  booktitle={Proceedings of the Computer Vision and Pattern Recognition Conference},
  pages={13294--13304},
  year={2025}
}

@article{zhang2025easycontrol,
  title={Easycontrol: Adding efficient and flexible control for diffusion transformer},
  author={Zhang, Yuxuan and Yuan, Yirui and Song, Yiren and Wang, Haofan and Liu, Jiaming},
  journal={arXiv preprint arXiv:2503.07027},
  year={2025}
}

@inproceedings{shin2025large,
  title={Large-scale text-to-image model with inpainting is a zero-shot subject-driven image generator},
  author={Shin, Chaehun and Choi, Jooyoung and Kim, Heeseung and Yoon, Sungroh},
  booktitle={Proceedings of the Computer Vision and Pattern Recognition Conference},
  pages={7986--7996},
  year={2025}
}

@inproceedings{zeng2024jedi,
  title={Jedi: Joint-image diffusion models for finetuning-free personalized text-to-image generation},
  author={Zeng, Yu and Patel, Vishal M and Wang, Haochen and Huang, Xun and Wang, Ting-Chun and Liu, Ming-Yu and Balaji, Yogesh},
  booktitle={Proceedings of the IEEE/CVF Conference on Computer Vision and Pattern Recognition},
  pages={6786--6795},
  year={2024}
}

@inproceedings{
  wang2025msdiffusion,
  title={{MS}-Diffusion: Multi-subject Zero-shot Image Personalization with Layout Guidance},
  author={Xierui Wang and Siming Fu and Qihan Huang and Wanggui He and Hao Jiang},
  booktitle={The Thirteenth International Conference on Learning Representations},
  year={2025},
  url={https://openreview.net/forum?id=PJqP0wyQek}
}

@inproceedings{ma2024subject,
  title={Subject-diffusion: Open domain personalized text-to-image generation without test-time fine-tuning},
  author={Ma, Jian and Liang, Junhao and Chen, Chen and Lu, Haonan},
  booktitle={ACM SIGGRAPH 2024 Conference Papers},
  pages={1--12},
  year={2024}
}

@inproceedings{
pan2024kosmosg,
title={Kosmos-G: Generating Images in Context with Multimodal Large Language Models},
author={Xichen Pan and Li Dong and Shaohan Huang and Zhiliang Peng and Wenhu Chen and Furu Wei},
booktitle={The Twelfth International Conference on Learning Representations},
year={2024},
url={https://openreview.net/forum?id=he6mX9LTyE}
}

@article{
patel2024eclipse,
title={\ensuremath{\lambda}-{ECLIPSE}: Multi-Concept Personalized Text-to-Image Diffusion Models by Leveraging {CLIP} Latent Space},
author={Maitreya Patel and Sangmin Jung and Chitta Baral and Yezhou Yang},
journal={Transactions on Machine Learning Research},
issn={2835-8856},
year={2024},
url={https://openreview.net/forum?id=7q5UewlAdM},
note={}
}

@inproceedings{wei2023elite,
  title={Elite: Encoding visual concepts into textual embeddings for customized text-to-image generation},
  author={Wei, Yuxiang and Zhang, Yabo and Ji, Zhilong and Bai, Jinfeng and Zhang, Lei and Zuo, Wangmeng},
  booktitle={Proceedings of the IEEE/CVF International Conference on Computer Vision},
  pages={15943--15953},
  year={2023}
}

@article{van2016conditional,
  title={Conditional image generation with pixelcnn decoders},
  author={Van den Oord, Aaron and Kalchbrenner, Nal and Espeholt, Lasse and Vinyals, Oriol and Graves, Alex and others},
  journal={Advances in neural information processing systems},
  volume={29},
  year={2016}
}

@inproceedings{
salimans2017pixelcnn,
title={Pixel{CNN}++: Improving the Pixel{CNN} with Discretized Logistic Mixture Likelihood and Other Modifications},
author={Tim Salimans and Andrej Karpathy and Xi Chen and Diederik P. Kingma},
booktitle={International Conference on Learning Representations},
year={2017},
url={https://openreview.net/forum?id=BJrFC6ceg}
}

@article{
yu2022scaling,
title={Scaling Autoregressive Models for Content-Rich Text-to-Image Generation},
author={Jiahui Yu and Yuanzhong Xu and Jing Yu Koh and Thang Luong and Gunjan Baid and Zirui Wang and Vijay Vasudevan and Alexander Ku and Yinfei Yang and Burcu Karagol Ayan and Ben Hutchinson and Wei Han and Zarana Parekh and Xin Li and Han Zhang and Jason Baldridge and Yonghui Wu},
journal={Transactions on Machine Learning Research},
issn={2835-8856},
year={2022},
url={https://openreview.net/forum?id=AFDcYJKhND},
note={Featured Certification}
}

@inproceedings{ControlAR,
      title={ControlAR: Controllable Image Generation with Autoregressive Models}, 
      author={Li, Zongming and Cheng, Tianheng and Chen, Shoufa and Sun, Peize and Shen, Haocheng and Ran, Longjin and Chen, Xiaoxin and Liu, Wenyu and Wang, Xinggang},
      booktitle={International Conference on Learning Representations},
      year={2025}
}

@inproceedings{
mentzer2024finite,
title={Finite Scalar Quantization: {VQ}-{VAE} Made Simple},
author={Fabian Mentzer and David Minnen and Eirikur Agustsson and Michael Tschannen},
booktitle={The Twelfth International Conference on Learning Representations},
year={2024},
url={https://openreview.net/forum?id=8ishA3LxN8}
}

@inproceedings{
yu2022vectorquantized,
title={Vector-quantized Image Modeling with Improved {VQGAN}},
author={Jiahui Yu and Xin Li and Jing Yu Koh and Han Zhang and Ruoming Pang and James Qin and Alexander Ku and Yuanzhong Xu and Jason Baldridge and Yonghui Wu},
booktitle={International Conference on Learning Representations},
year={2022},
url={https://openreview.net/forum?id=pfNyExj7z2}
}

@article{li2024autoregressive,
  title={Autoregressive image generation without vector quantization},
  author={Li, Tianhong and Tian, Yonglong and Li, He and Deng, Mingyang and He, Kaiming},
  journal={Advances in Neural Information Processing Systems},
  volume={37},
  pages={56424--56445},
  year={2024}
}

@inproceedings{
fan2025fluid,
title={Fluid: Scaling Autoregressive Text-to-image Generative Models with Continuous Tokens},
author={Lijie Fan and Tianhong Li and Siyang Qin and Yuanzhen Li and Chen Sun and Michael Rubinstein and Deqing Sun and Kaiming He and Yonglong Tian},
booktitle={The Thirteenth International Conference on Learning Representations},
year={2025},
url={https://openreview.net/forum?id=jQP5o1VAVc}
}

@article{wu2024janus,
  title={Janus: Decoupling visual encoding for unified multimodal understanding and generation},
  author={Wu, Chengyue and Chen, Xiaokang and Wu, Zhiyu and Ma, Yiyang and Liu, Xingchao and Pan, Zizheng and Liu, Wen and Xie, Zhenda and Yu, Xingkai and Ruan, Chong and others},
  journal={arXiv preprint arXiv:2410.13848},
  year={2024}
}

@article{yao2024car,
  title={Car: Controllable autoregressive modeling for visual generation},
  author={Yao, Ziyu and Li, Jialin and Zhou, Yifeng and Liu, Yong and Jiang, Xi and Wang, Chengjie and Zheng, Feng and Zou, Yuexian and Li, Lei},
  journal={arXiv preprint arXiv:2410.04671},
  year={2024}
}

@article{li2024controlvar,
  title={Controlvar: Exploring controllable visual autoregressive modeling},
  author={Li, Xiang and Qiu, Kai and Chen, Hao and Kuen, Jason and Lin, Zhe and Singh, Rita and Raj, Bhiksha},
  journal={arXiv preprint arXiv:2406.09750},
  year={2024}
}

@inproceedings{
podell2024sdxl,
title={{SDXL}: Improving Latent Diffusion Models for High-Resolution Image Synthesis},
author={Dustin Podell and Zion English and Kyle Lacey and Andreas Blattmann and Tim Dockhorn and Jonas M{\"u}ller and Joe Penna and Robin Rombach},
booktitle={The Twelfth International Conference on Learning Representations},
year={2024},
url={https://openreview.net/forum?id=di52zR8xgf}
}

@article{hu2022lora,
  title={Lora: Low-rank adaptation of large language models.},
  author={Hu, Edward J and Shen, Yelong and Wallis, Phillip and Allen-Zhu, Zeyuan and Li, Yuanzhi and Wang, Shean and Wang, Lu and Chen, Weizhu and others},
  journal={ICLR},
  volume={1},
  number={2},
  pages={3},
  year={2022}
}

@article{
oquab2024dinov,
title={{DINO}v2: Learning Robust Visual Features without Supervision},
author={Maxime Oquab and Timoth{\'e}e Darcet and Th{\'e}o Moutakanni and Huy V. Vo and Marc Szafraniec and Vasil Khalidov and Pierre Fernandez and Daniel HAZIZA and Francisco Massa and Alaaeldin El-Nouby and Mido Assran and Nicolas Ballas and Wojciech Galuba and Russell Howes and Po-Yao Huang and Shang-Wen Li and Ishan Misra and Michael Rabbat and Vasu Sharma and Gabriel Synnaeve and Hu Xu and Herve Jegou and Julien Mairal and Patrick Labatut and Armand Joulin and Piotr Bojanowski},
journal={Transactions on Machine Learning Research},
issn={2835-8856},
year={2024},
url={https://openreview.net/forum?id=a68SUt6zFt},
note={Featured Certification}
}

@inproceedings{liu2024grounding,
  title={Grounding dino: Marrying dino with grounded pre-training for open-set object detection},
  author={Liu, Shilong and Zeng, Zhaoyang and Ren, Tianhe and Li, Feng and Zhang, Hao and Yang, Jie and Jiang, Qing and Li, Chunyuan and Yang, Jianwei and Su, Hang and others},
  booktitle={European conference on computer vision},
  pages={38--55},
  year={2024},
  organization={Springer}
}

@inproceedings{chang2023muse,
  title={Muse: Text-To-Image Generation via Masked Generative Transformers},
  author={Chang, Huiwen and Zhang, Han and Barber, Jarred and Maschinot, Aaron and Lezama, Jose and Jiang, Lu and Yang, Ming-Hsuan and Murphy, Kevin Patrick and Freeman, William T and Rubinstein, Michael and others},
  booktitle={International Conference on Machine Learning},
  pages={4055--4075},
  year={2023},
  organization={PMLR}
}

@inproceedings{
ho2021classifierfree,
title={Classifier-Free Diffusion Guidance},
author={Jonathan Ho and Tim Salimans},
booktitle={NeurIPS 2021 Workshop on Deep Generative Models and Downstream Applications},
year={2021},
url={https://openreview.net/forum?id=qw8AKxfYbI}
}

@article{Qwen2.5-VL,
  title={Qwen2.5-VL Technical Report},
  author={Bai, Shuai and Chen, Keqin and Liu, Xuejing and Wang, Jialin and Ge, Wenbin and Song, Sibo and Dang, Kai and Wang, Peng and Wang, Shijie and Tang, Jun and Zhong, Humen and Zhu, Yuanzhi and Yang, Mingkun and Li, Zhaohai and Wan, Jianqiang and Wang, Pengfei and Ding, Wei and Fu, Zheren and Xu, Yiheng and Ye, Jiabo and Zhang, Xi and Xie, Tianbao and Cheng, Zesen and Zhang, Hang and Yang, Zhibo and Xu, Haiyang and Lin, Junyang},
  journal={arXiv preprint arXiv:2502.13923},
  year={2025}
}

@inproceedings{
chen2023reimagen,
title={Re-Imagen: Retrieval-Augmented Text-to-Image Generator},
author={Wenhu Chen and Hexiang Hu and Chitwan Saharia and William W. Cohen},
booktitle={The Eleventh International Conference on Learning Representations },
year={2023},
url={https://openreview.net/forum?id=XSEBx0iSjFQ}
}

@article{chung2025fine,
  title={Fine-Tuning Visual Autoregressive Models for Subject-Driven Generation},
  author={Chung, Jiwoo and Hyun, Sangeek and Kim, Hyunjun and Koh, Eunseo and Lee, MinKyu and Heo, Jae-Pil},
  journal={arXiv preprint arXiv:2504.02612},
  year={2025}
}

@article{tschannen2025siglip,
  title={Siglip 2: Multilingual vision-language encoders with improved semantic understanding, localization, and dense features},
  author={Tschannen, Michael and Gritsenko, Alexey and Wang, Xiao and Naeem, Muhammad Ferjad and Alabdulmohsin, Ibrahim and Parthasarathy, Nikhil and Evans, Talfan and Beyer, Lucas and Xia, Ye and Mustafa, Basil and others},
  journal={arXiv preprint arXiv:2502.14786},
  year={2025}
}
\bibliographystyle{main}

\newpage
\section{Appendix}

\subsection{Dataset}
We train EchoGen on the filtered combination of the Subjects200K and UNO-1M synthetic datasets.
The Subjects200K dataset\footnote{\url{https://huggingface.co/datasets/Yuanshi/Subjects200K}}, introduced by~\citep{tan2025ominicontrol}, contains approximately 256,000 triplets, each comprising a reference image, a text prompt, and a corresponding generated target image. It is specifically established for subject-driven generation. The dataset is constructed by first using GPT-4o to produce over 30,000 diverse subject descriptions that each description represents the same subject across multiple scenes. These descriptions are then reformulated into structured prompts, each specifying a single subject in two different scenes, which are fed into FLUX.1-dev text-to-image model to synthesize paired images. Finally, GPT-4o filters the generated pairs to ensure subject consistency and overall image quality. All images have the resolution above 500 pixels, providing sufficient detail for training.

The UNO-1M dataset\footnote{\url{https://huggingface.co/datasets/bytedance-research/UNO-1M}}~\citep{wu2025less} comprises approximately 1M data triplets and is built in a similar manner. It leverages the LLM to generate diverse subject instances and scenes guided by a taxonomy tree derived from the 365 general classes of Object-365, and then employs FLUX.1 model to synthesize image pairs. After that, DINO-v2 and Vision–Language Models (VLMs) are further used to score and filter the generated data. To obtain a high-quality corpus, we use their single-object subset and additionally filter it based on the VLM scores, resulting in approximately 394,000 triplets.

\subsection{Pseudo-Code of the EchoGen Block}
We provide a PyTorch-style pseudocode for our EchoGen Block in \Cref{algo:echogen} to facilitate reproducibility and clarity.
\begin{algorithm}[h]
\caption{EchoGen Block: PyTorch-like Pseudo-code}
\label{algo:echogen}
\algcomment{
\textbf{Pseudo-code illustrating the EchoGen Block}.
Here, $x$ denotes the image token sequence, and the generation process is conditioned on the semantic feature $c_s$ and detailed content feature $c_c$ extracted from the reference image, along with the text embedding $c_t$.
}
\definecolor{codeblue}{rgb}{0.25,0.5,0.5}
\definecolor{codekw}{rgb}{0.85, 0.18, 0.50}
\begin{lstlisting}[language=python]
class EchoGenBlock(nn.Module)
    def __init__(dim, mask):
        # Multi-modal attention QKV projectors for the image token sequence
        self.qkv_mm = nn.Linear(dim, 3*dim)
        # Multi-modal attention QKV projectors for the detailed content feature 
        self.qkv_mm_c = nn.Linear(dim, 3*dim)
        self.mask = mask

        # Cross attention query projectors for the image token sequence
        self.q_ca = nn.Linear(dim, dim)
        # Cross attention KV projectors for the semantic feature
        self.kv_ca_s = nn.Linear(dim, 2*dim)
        # Cross attention KV projectors for the text embedding
        self.kv_ca_t = nn.Linear(dim, 2*dim)

        # FFN for the image token sequence
        self.ffn = MLP(dim)
        # FFN for the detailed content feature 
        self.ffn_c = MLP(dim)
        
    def forward(self, x, cc, cs, ct):
        # Multi-modal attention
        q, k, v = self.qkv_mm(x)
        qc, kc, vc = self.qkv_mm(cc)

        q = torch.concat((q, qc))
        k = torch.concat((k, kc))
        v = torch.concat((v, vc))
        x, cc = attention(q, k, v, self.mask)

        # Cross attention
        q = self.q_ca(x)
        ks, vs = self.kv_ca_s(cs)
        kt, vt = self.kv_ca_t(ct)
        
        k = torch.concat((ks, kt))
        v = torch.concat((vs, vt))
        x = attention(q, k, v, mask=None)

        # Feed-forward network
        x = self.ffn(x)
        cc = self.ffn_c(cc)
        
        return x, cc, cs, ct
\end{lstlisting}
\end{algorithm}

\subsection{Implementation Details}
\subsubsection{Training Details}\label{sec:traindetail}
\textbf{Data Preprocssing}. All training images undergo a standardized pre-processing pipeline. Initially, images are resized so that their shorter edge matches the model's target resolution, which is 256 pixels for EchoGen-0.1B and 1024 pixels for EchoGen-2B, followed by a central crop to achieve a square aspect ratio. 

To maintain data quality for high-resolution image generation within the EchoGen-2B model, we circumvent the quality degradation induced by naive bilinear upsampling. Instead of using simple bilinear scaling on images smaller than 1024 pixels, we integrate a super-resolution step. Specifically, we leverage the PiSA-SR model~\citep{sun2025pixel} to upscale these images, a method chosen to preserve fine-grained textures and prevent the introduction of common interpolation artifacts. This ensures that the model is trained exclusively on high-quality and high-resolution exemplars.

\textbf{Training Hyper-parameters}.
We follow the Infinity-2B standard training recipe~\citep{han2025infinity}, and the detailed hyperparameter configurations used to train our EchoGen are provided in \Cref{tab:hyper}. To mitigate error accumulation, as mentioned in Infinity, we employ bitwise self-correction by randomly flipping bits in the input sequence with a probability of 0.3. To improve robustness against variations in instruction length, prompts are randomly truncated to a single sentence with a probability of 0.5 during training. The EchoGen models are trained using 32 H20 GPUs, requiring 2 weeks for the longest schedule (training our EchoGen-2B model for 20 epochs).

\begin{table*}[t]
\centering
\begin{tabular}{ll}
\toprule
Config & value \\
\midrule
Bitwise Self-correction Flip Ratio & 0.3 \\
Bitwise Self-correction Apply Layers & 13 \\
Dynamic Truncate Prompt Ratio & 0.5 \\ 
\midrule
Infinity Image Encoder Channel & 16(0.1B) / 32(2B) \\
\midrule
Text Encoder & Flan-t5-xl \\
Text Embedding Channels & 2048 \\
Maximum Text Tokens Length & 512\\
\midrule
Semantic Image Encoder & DINO-v2-Base\\
Semantic Feature Channels & 768 \\
Semantic Downsample ratio & 14 \\
\midrule
Content Image Encoder & FLUX.1-dev VAE\\
Content Feature Channels & 16 \\
Content Downsample ratio & 8 \\
\midrule
Reweight Loss by Scale & True\\
\midrule
Gradient clipping by norm & 5.0 \\
Optimizer & Adamw \\
Beta1 & 0.9 \\
Beta2 & 0.97 \\
Decay & 0 \\
Base Learning rate & 3e-5 \\
Multi-Modal Modules Learning rate & 3e-6 \\
Learning rate warmup iterations & 0 \\
\midrule
Training epochs & 20 \\
Total Batchsize & 128 \\
GPU & H20 \\
\bottomrule
\end{tabular}
\caption{
\textbf{Detailed hyper-parameters for training our EchoGen}. 
}
\label{tab:hyper}
\end{table*}

\subsubsection{Evaluation Details}
For our quantitative evaluation, we utilize the DreamBench dataset~\citep{ruiz2023dreambooth}. The dataset comprises 30 distinct subjects, categorized into 9 animate pets (cats and dogs) and 21 diverse inanimate objects (\textit{e.g.}, toys, sunglasses, backpacks). Each subject is associated with 25 textual prompts specifically designed to test the model's abilities in recontextualization, property modification, and accessorization. Our data preparation protocol is adapted from~\citep{pan2024kosmosg}, which involves selecting a single reference image per subject and augmenting its subject identity phrase with descriptive keywords. 
The correspondence between the DreamBench dataset directory name and the augmented subject description is summarized as follows:
\begin{itemize}
\item backpack, backpack
\item backpack\_dog, dog shaped backpack
\item bear\_plushie, bear plushie
\item can, 'Transatlantic IPA' can
\item candle, jar candle
\item cat, tabby cat
\item cat2, grey cat
\item clock, number '3' clock
\item colorful\_sneaker, colorful sneaker
\item dog1, fluffy dog
\item dog2, fluffy dog
\item dog3, curly-haired dog
\item dog5, long-haired dog
\item dog6, puppy
\item dog7, dog
\item dog8, dog
\item duck\_toy, duck toy
\item fancy\_boot, fringed cream boot
\item grey\_sloth\_plushie, grey sloth plushie
\item monster\_toy, monster toy
\item pink\_sunglasses, sunglasses
\item poop\_emoji, poop-emoji shaped toy
\item rc\_car, car toy
\item red\_cartoon, cartoon character
\item robot\_toy, robot toy
\item shiny\_sneaker, sneaker
\item teapot, clay teapot
\item vase, tall vase
\item wolf\_plushie, wolf plushie
\end{itemize}

We compute DINO and CLIP-I scores by comparing each generated image with its corresponding single reference image in our main experiments and ablation studies. Note that some methods~\citep{li2023blip} especially test-time fine-tuning methods~\citep{ruiz2023dreambooth} instead compute DINO and CLIP-I by comparing a generated image against all images of the same entity in DreamBench.
To enable a more fair comparison under both protocols, we evaluate EchoGen and several leading baselines using unified implementation and report all results in~\Cref{sec:more_abla}.

To augment the diversity and rigor of our human evaluation, we incorporate a curated set of instances from the DreamBench++ benchmark. DreamBench++ includes 150 subjects, each paired with nine prompts.

\subsection{More ablation studies}\label{sec:more_abla}
In this section, we present additional ablation studies to analyze the individual components of EchoGen. These ablation studies are also conducted based on EchoGen-0.1B model with fair training settings.

\begin{wraptable}{r}{0.5\textwidth}
\centering
\renewcommand{\arraystretch}{0.95}
\vspace{-0.4cm}
\begin{tabular}{cccc}
\toprule
 Exp. & DINO$\uparrow$ & CLIP-I$\uparrow$ & CLIP-T$\uparrow$ \\
\midrule
Content & 0.663 & 0.795 & \textbf{0.322} \\
Semantic & \textbf{0.672} & \textbf{0.806} & 0.321 \\
\bottomrule
\end{tabular}
\caption{ \textbf{Analysis of different information types prepended to the Image Token.} Content represents prepending the global content feature, while Semantic denotes prepending the global semantic feature.
}
\label{tab:global}
\end{wraptable}
\noindent\textbf{Importance of injecting global semantic information}. 
Injecting global semantic information serves as a prepended condition, to ensure global structural coherence during generation. Our ablation study in \Cref{tab:prefix} confirms the significant benefits of incorporating global semantic features. Moreover, we conduct a targeted experiment comparing the injection of global semantic versus content features to the Image Token. As shown in \Cref{tab:global}, the results clearly indicate that prepending semantic features rather than content information into the image token significantly enhances the subject fidelity. This confirms that our choice to inject global semantic guidance into image tokens is both effective and well-justified.

\begin{wraptable}{r}{0.5\linewidth} 
    \vspace{-0.4cm}
    \centering
    \renewcommand{\arraystretch}{0.95}
    \begin{tabular}{cccc}
    \toprule
     Exp. & DINO$\uparrow$ & CLIP-I$\uparrow$ & CLIP-T$\uparrow$ \\
    \midrule
    w/o SS & 0.737 & 0.829 & 0.324 \\
    Shift & 0.739 & 0.833 & 0.321 \\
    Enlarge & 0.735 & 0.831 & 0.321 \\
    w/ SS & \textbf{0.755} & \textbf{0.835} & \textbf{0.325} \\
    \bottomrule
    \end{tabular}
    \caption{ \textbf{Analysis of the sensitivity of EchoGen to the segmentation quality.} SS denotes the subject segmentation; Enlarge and Shift denote enlarging and shifting bounding box, respectively.
    }
    \label{tab:segmentation}
\end{wraptable}
\textbf{Robustness to segmentation quality}.
We further analyze the sensitivity of our method to the quality of segmentation, we conduct an ablation study on about EchoGen's robustness to segmentation quality during inference. Specifically, to simulate segmentation imperfections, we design three variants to simulate disturbances and compare with employing subject segmentation without imperfection during inference: 1. Enlarging Bounding Box: enlarging the subject's bounding box by 10\%; 2. Shifting Bounding Box: shifting the bounding box by 10\%; 3. No Segmentation: completely removing the subject segmentation step. As shown in \Cref{tab:segmentation}, our model exhibits remarkable robustness as its performance degrades only slightly under these disturbances. Moreover, our model still produces strong results even without any segmentation, demonstrating its powerful generalization capability. In summary, EchoGen is highly robust to imperfect segmentation.

\begin{table*}[t]
\centering
\setlength{\tabcolsep}{3.6pt}

\begin{subtable}[t]{\textwidth}
\centering
\begin{tabular}{llcccc}
\toprule
Method & Base Model & DINO$\uparrow$ & CLIP-I$\uparrow$ & CLIP-T$\uparrow$ & Latency$\downarrow$\\
\midrule
IP-Adapter~\citep{ye2023ip} & SDXL & 0.713 & 0.851 & 0.293 & 16.9s\\
OminiControl~\citep{tan2025ominicontrol} & FLUX.1-dev  & 0.644 & 0.800 & 0.323 & 27.5s\\
EasyControl~\citep{zhang2025easycontrol} & FLUX.1-dev & 0.666 & 0.804 & 0.330 & 25.4s\\
\midrule
EchoGen-2B & Infinity-2B &  0.755 & 0.835 & 0.325 & 5.2s\\
\bottomrule
\end{tabular}
\caption{Multi-reference scoring (compare against all references of the same entity).}
\label{tab:metrics_multi}
\end{subtable}

\begin{subtable}[t]{\textwidth}
\centering
\begin{tabular}{llcccc}
\toprule
Method & Base Model & DINO$\uparrow$ & CLIP-I$\uparrow$ & CLIP-T$\uparrow$ & Latency$\downarrow$\\
\midrule
IP-Adapter~\citep{ye2023ip} & SDXL & 0.626 & 0.807 & 0.295 & 16.9s\\
OminiControl~\citep{tan2025ominicontrol} & FLUX.1-dev  & 0.575 & 0.771 & 0.323 & 27.5s\\
EasyControl~\citep{zhang2025easycontrol} & FLUX.1-dev & 0.600 & 0.773 & 0.330 & 47.6s\\
\midrule
EchoGen-2B & Infinity-2B & 0.663 & 0.802 & 0.325 & 5.2s\\
\bottomrule
\end{tabular}
\caption{Single-reference scoring (compare against the corresponding reference image).}
\label{tab:metrics_single}
\end{subtable}

\caption{\textbf{Quantitative comparisons on DreamBench~\citep{ruiz2023dreambooth} under two evaluation protocols.} Baselines are reproduced from official repositories and evaluated with the same code. EchoGen achieves strong performance with lower latency.}
\label{tab:metrics}
\end{table*}

\textbf{Different evaluation protocols.}
We compare our model with several leading baselines under two different evaluation protocols: (i) each generated image is compared against the single corresponding reference image, and (ii) each generated image is compared against all reference images of the same entity in DreamBench. We reproduce the baselines using their official repositories and evaluate all methods with the same evaluation code.
As shown in~\Cref{tab:metrics}, our method achieves comparable or superior subject preservation, as measured by DINO and CLIP-I, as well as comparable text alignment (CLIP-T), while maintaining faster sampling speed.

\subsection{More visualization results}
\begin{figure*}[t]
    \centering
    \includegraphics[width=0.99\linewidth]{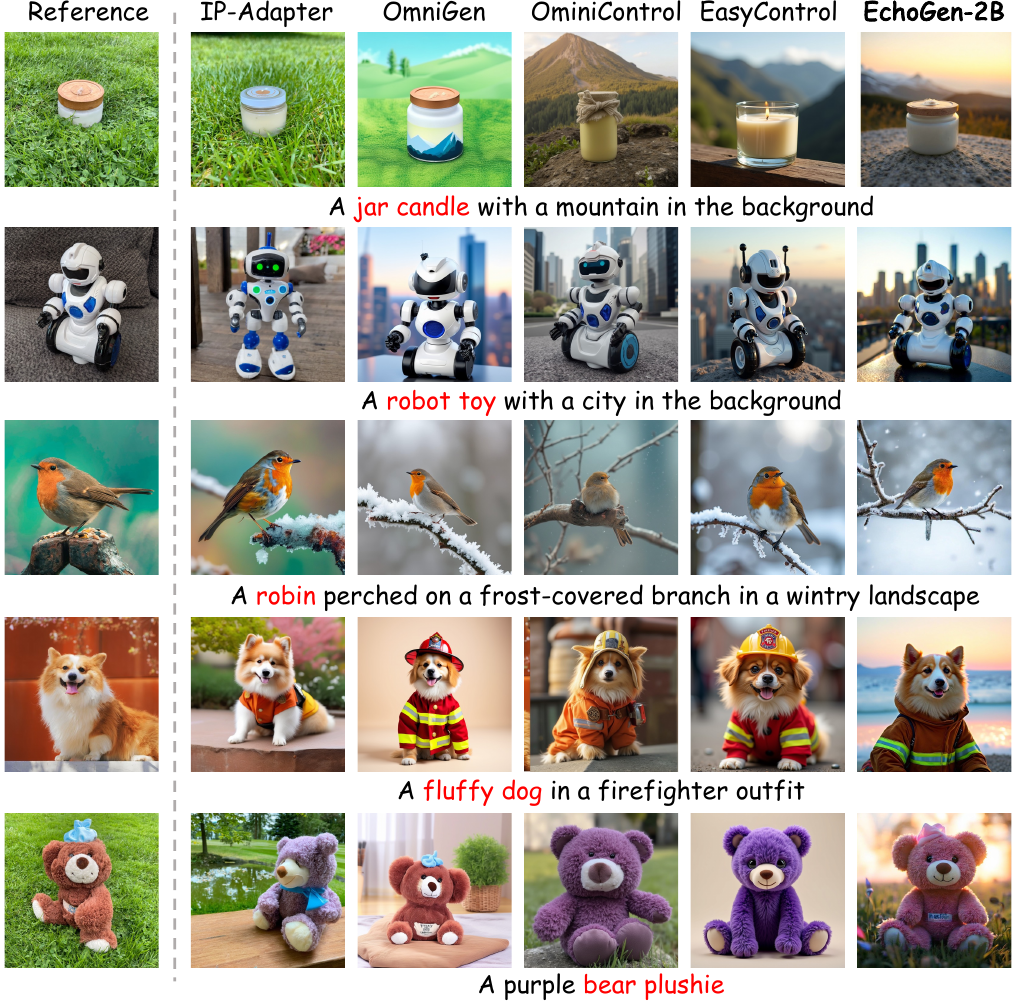}
    \caption{
    \textbf{Additional qualitative comparisons between our EchoGen model and competing methods.}}
    \label{fig:additional}
\end{figure*}

\begin{wrapfigure}{r}{0.5\linewidth}
        \centering
        \vspace{-0.8cm}
    \includegraphics[width=0.99\linewidth]{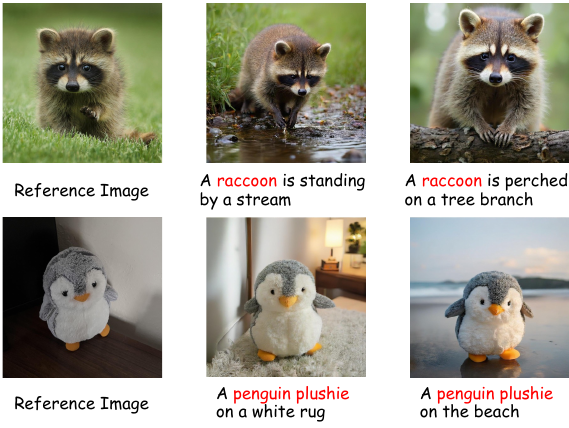}
    \vspace{-0.8cm}
    \caption{
    \textbf{More visualization of EchoGen-2B on real-world subject personalization.}}
    \label{fig:real-world}
    \vspace{-0.5cm}
\end{wrapfigure}
We further showcase additional qualitative results on DreamBench in \Cref{fig:additional}. Moreover, we provide additional visual results from the EchoGen-2B model on real-world subject personalization in \Cref{fig:real-world}. These results demonstrate that our model, trained exclusively on a filtered combination of the large-scale synthetic Subjects200K~\citep{tan2025ominicontrol} and UNO-1M~\citep{wu2025less} datasets, our model exhibits strong generalization to real-world scenarios, including the generation of live animals and diverse objects under complex conditions. EchoGen-2B consistently maintains high subject fidelity and strong text alignment during these real-world personalization tasks, demonstrating the effectiveness of our training strategy and the proposed dual-path semantic-content injection design.

\subsection{Limitation \& Failure case analysis}
Our EchoGen takes a new step toward VAR-based feed-forward subject-driven generation to inherit the strong capability of next-scale prediction and bidirectional modeling within scales. However, we know that the feed-forward subject-driven image generation is highly dependent on the capability of base models. 
The performance of our EchoGen models is fundamentally dependent upon the capability of the base models Infinity-0.1B and Infinity-2B. The Infinity-2B architecture still exhibits a performance gap compared to state-of-the-art generation models such as Stable-Diffusion 3 and FLUX, particularly in generating high-fidelity details. 
This inherited constraint limits EchoGen's efficacy in resolving fine-grained features, such as the faithful rendering of facial characteristics, the synthesis of coherent text, and the reproduction of intricate material textures.
Due to significant GPU computational and temporal constraints, our experiments are confined to these specific backbones, precluding an empirical investigation of larger models such as Infinity-8B. We hypothesize that migrating the EchoGen architecture to a more potent VAR foundation model would unlock substantial performance gains. 

Additionally, the DINOv2 vision encoder operates on relatively low-resolution inputs (\textit{e.g.}, 224×224), which limits its ability to capture fine-grained appearance cues and tiny textual elements. We believe seeking an effective high-resolution semantic encoder presents a promising avenue for further improvement in complex applications.

\begin{figure*}[t]
    \centering
    \includegraphics[width=0.99\linewidth]{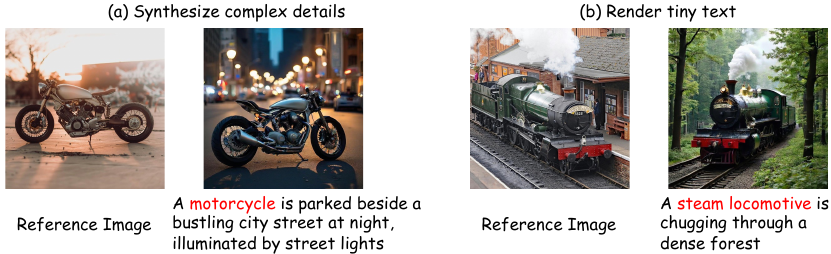}
    \caption{
    \textbf{Failure cases generated by EchoGen.}}
    \label{fig:failure}
\end{figure*}

Due to the aforementioned limitations, as illustrated in \Cref{fig:failure}, the model exhibits reduced reliability on subjects that have highly intricate structures or in scenarios requiring precise text rendering. We will continue to investigate and address these challenges in future work.
\end{document}